%% file: _main.tex
\documentclass[11pt]{article}
\usepackage{acl}
\usepackage{times}
\usepackage{latexsym}
\usepackage[T1]{fontenc}
\usepackage[utf8]{inputenc}
\usepackage{microtype}
\usepackage{inconsolata}
\usepackage{graphicx}
\usepackage{subcaption}
\usepackage{pifont}
\usepackage{xcolor}
\newcommand{\cmark}{\textcolor{green}{\ding{51}}} 
\newcommand{\xmark}{\textcolor{red}{\ding{55}}}   

\title{When Verification Hurts: Asymmetric Effects of Multi-Agent Feedback in Logic Proof Tutoring}

\author{
Tahreem Yasir, Sutapa Dey Tithi, Benyamin Tabarsi, Dmitri Droujkov, Sam Gilson \\
Yasitha Rajapaksha, Xiaoyi Tian, DongKuan (DK) Xu, Tiffany Barnes \\
North Carolina State University \\
Raleigh, NC, USA \\
\texttt{\{tyasir, stithi, btaghiz, ddroujk, sagilson, yrajapa, xtian9, dxu27, tmbarnes\}@ncsu.edu}
\AND
Arun Ramesh \\
\texttt{rarunkumar15112k@gmail.com}
}

\usepackage{tikz}
\usetikzlibrary{arrows.meta, positioning, shapes, calc}
\usepackage{listings}
\usepackage{booktabs}
\usepackage{multirow}
\usepackage{makecell}
\usepackage{amssymb}
\usepackage{amsmath} 
\usepackage{float}
\usepackage{ragged2e}
\usepackage[utf8]{inputenc}
\usepackage{textcomp} 
\usepackage{enumitem}
\usepackage[most]{tcolorbox}
\usepackage{hyperref}

\begin{document}

\maketitle
\input{latex/framework_Image}
\input{latex/0_abstract}
\input{latex/1_introduction}
\input{latex/2_related_work}

\input{latex/3_dataset}

\input{latex/4_method}

\input{latex/5_evaluation_metrics}

\input{latex/6_experiment_setup}
\input{latex/7_results}

\input{latex/8_conclusions}

\input{latex/limit_future}

\bibliography{custom}

\input{latex/appendix}

\end{document}

%% file: latex/framework_Image.tex
\begin{figure*}[ht]
    \centering

    \includegraphics[width=0.65\textheight]{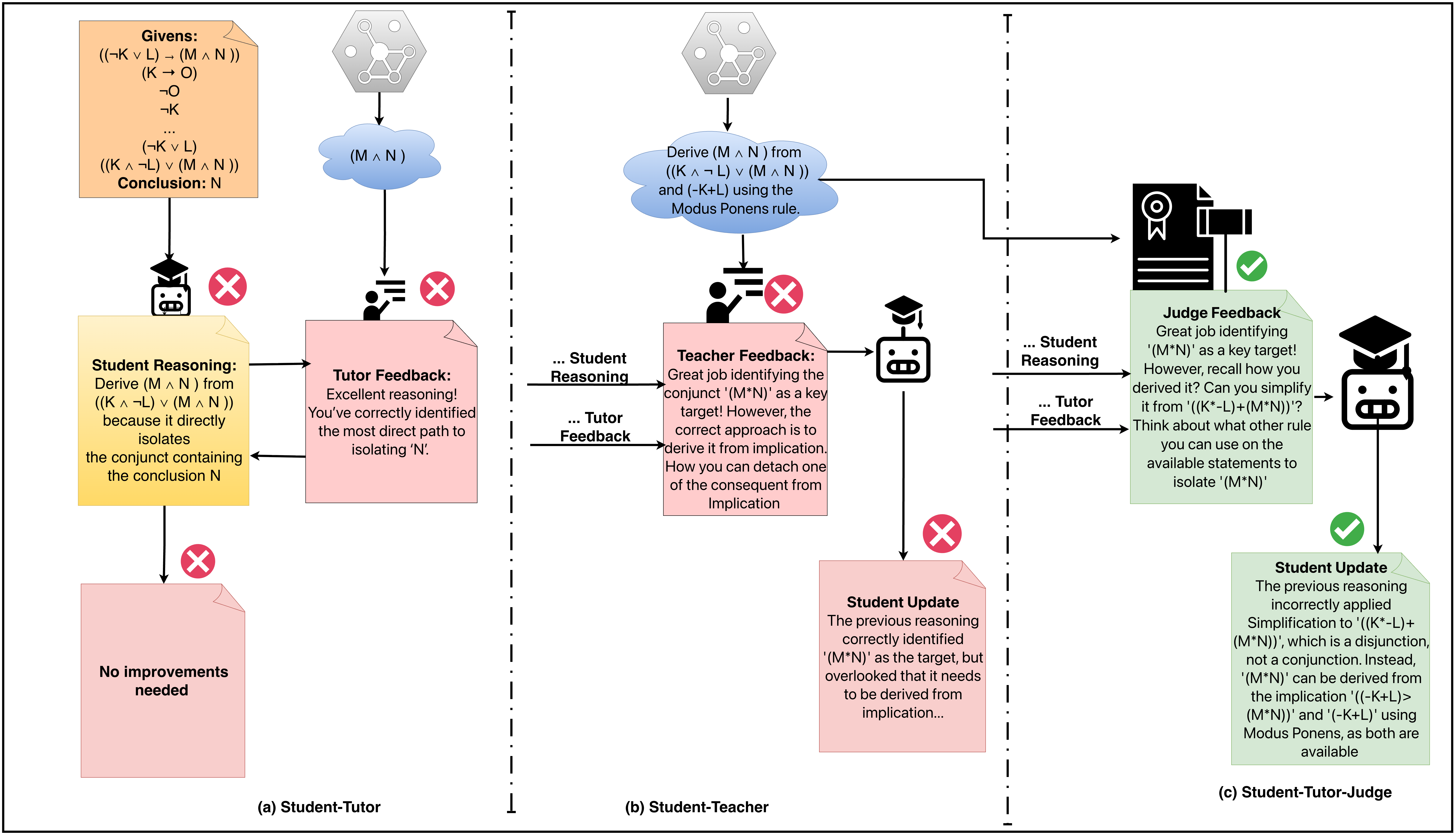}
    \caption{Multi-agent Tutoring with KG-grounded varying hints
    \textbf{(a)} Student solution is partially correct (using incorrect rule) \textcolor{red}{Tutor} echoes student's incorrect rule, validates partial correctness \textbf{(b)}\textcolor{red}{Teacher} reveals solutions prematurely and students anchors \textbf{(c)} \textcolor{blue}{Judge} correctly identifies the error in students reasoning (using incorrect rules and statements for derivations)}
    \label{fig:framework}
    
\end{figure*}

%% file: latex/0_abstract.tex
\begin{abstract}
Large language models (LLMs) are increasingly used for automated tutoring, 
but their reliability in structured symbolic domains remains unclear. 
We 
study step-level feedback for propositional logic proofs, which require 
precise symbolic reasoning aligned with a learner's current proof state. We introduce a knowledge-graph–grounded benchmark of 516 unique proof states 
with step-level annotations and difficulty metrics. Unlike prior tutoring 
evaluations that rely on model self-assessment or binary correctness, our 
framework enables fine-grained analysis of feedback quality against 
verified solution paths. We evaluate three role-specialized pipelines with varying solution access: Tutor 
(partial solution access), Teacher (full derivation access), and Judge 
(verification of Tutor feedback). Our results reveal a striking asymmetry: verification improves outcomes 
when upstream feedback is error-prone (<70\% accuracy), but 
\emph{degrades} performance by 4–6pp through over-specification when 
feedback is already reliable (>85\%). Critically, we identify a shared 
complexity ceiling; no model or pipeline reliably succeeds on proof states 
exceeding complexity 4–5.
These findings challenge the assumption that adding verifiers or richer 
context universally improves tutoring, motivating adaptive, difficulty-aware 
architectures that route problems by estimated complexity and upstream 
reliability.
\end{abstract}

%% file: latex/1_introduction.tex
\section{Introduction}
Propositional logic is foundational to computer science education, underpinning formal reasoning in programming, algorithms, and digital systems \citep{mayer2025importance}. Yet mastering optimal proof strategies remains challenging for students due to limited access to timely and personalized tutoring \citep{inamdar2025logiclearner}. 


Intelligent tutoring systems (ITS) achieve learning gains comparable to human tutoring through \emph{step-level feedback}, guiding each inference rather than evaluating only final answers \citep{vanlehn2011effectiveness}. However, traditional ITS relies on template-based feedback that poorly adapts to diverse misconceptions \citep{zerkouk2025its_review}. Large language models (LLMs) offer scalable alternatives, yet their tendency to hallucinate risks incorrect feedback undermining learning \citep{macina2023mathdial}.

Effective step-level tutoring requires more than surface correctness: feedback that is partially correct or misaligned with a student's proof state can reinforce misconceptions and anchor flawed reasoning even when not strictly wrong \citep{vanlehn2011effectiveness}. Without external grounding, even capable LLMs risk generating feedback that appears correct but misguides student reasoning, motivating the need for verification mechanisms \citep{dhuliawala2024chain}.


Role-specialized feedback pipelines, where a dedicated verifier agent evaluates another agent's output, offer a potential safeguard distinct from self-correction \citep{huang2024selfcorrect}. Yet the conditions under which verification helps, or hurts, 
is not well studied.
A key factor also overlooked in prior work is the degree of
solution \emph{information access} for the
feedback agent. Greater 
access may enable more precise verification, but 
risks prematurely revealing answers and undermining 
scaffolding \citep{daheim2024stepwise}. Moreover, how 
information access shapes verification effectiveness, 
what failure modes emerge, and whether automated 
accuracy metrics capture the underlying pedagogical characteristics remain largely unexplored. We address these gap in our work with the following research questions:

\noindent\textit{\textbf{RQ1:}} How does the degree of solution information access shape step-level tutoring feedback effectiveness across models of varying capability? \\
\noindent \textit{\textbf{RQ2:}} Under what conditions does role-specialized verification improve tutoring feedback, and when does it degrade performance? \\
\noindent\textit{\textbf{RQ3:}} What systematic failure modes emerge across feedbacks in each pipelines, and whether automated metrics alone capture their pedagogical impact?

We answer these RQs through step-level feedback for 
propositional logic proofs, where students apply 
inference rules (e.g., Modus Ponens, Disjunctive 
Syllogism) to a set of premises (logical statements 
assumed to hold) to derive a target conclusion through 
intermediate derivations or steps. At each proposed step a tutoring agent must assess 
the derivation and provide scaffolded guidance 
without revealing the answer. For our investigation, we developed three role-specialized agents (Fig.~\ref{fig:framework}): Tutor (access only to the correct next step), Teacher (full solution access), and Judge (verification of Tutor feedback along with full solution access), operationalizing solution information access as a controlled variable. Unlike prior evaluations relying on binary correctness, we perform a fine-grained evaluation against a knowledge graph (KG), encoding the complete derivation space, enabling analysis of \emph{how} steps are derived, not only \emph{whether} they are valid. Our contributions are as follows:

\noindent\textbf{Logic Tutoring Benchmark:} A 
novel dataset of 516 real student propositional logic 
proof states with a KG-encoded multi-path solution 
space and instance-level difficulty metrics, 
enabling step-level evaluation beyond binary 
correctness and supporting the first controlled study to the best of our knowledge on information access effects in logic proof tutoring.

\noindent\textbf{Asymmetric Verification Effect:} 
Empirical evidence that the verification benefit is 
conditional on upstream feedback reliability, 
improving outcomes when feedback is 
error-prone, but degrading performance
through over-specification when feedback is already 
reliable.

\noindent\textbf{Failure Mode Taxonomy:} Four 
systematic failure modes, echo effects, partial 
correctness validation, premature revelation, and 
verification over-specification, confirmed by 
human evaluation across four pedagogical dimensions, 
demonstrating that automated accuracy metrics 
systematically overestimate feedback quality.



%% file: latex/2_related_work.tex
\section{Related Work} \label{sec:related}
\subsection{LLM-based scaffolding for tutoring systems}


Scaffolding theory \citep{vygotsky1978mind}  
underlies intelligent tutoring systems (ITS), where 
step-by-step hints must be specific enough to advance 
reasoning yet scaffolded enough to preserve productive 
struggle \citep{wood1976scaffolding, belland2017scaffolding}. This balance 
is especially difficult to achieve in structured symbolic domains where each inference step constrains all subsequent reasoning. 

LLMs show promise for generating adaptive feedback 
across diverse educational contexts, with prior work 
demonstrating gains in diagnosing student errors, and generating pedagogical feedback in college algebra tutors \citep{reddig2025generating}, fostering active reasoning and self-reflection in advanced computer science courses \citep{Kumar2024Supporting}, and supporting adaptive scaffolding in programming contexts \citep{scholz2025partnering}.
However, two persistent challenges limit their effectiveness in intelligent tutoring systems: LLMs often reveal answers instead of scaffolded guidance \citep{macina2023mathdial, macina2025mathtutorbench}, and their feedback frequently hallucinates without external grounding \citep{liu2025improving, jia2024assessing}. Moreover, LLMs cannot reliably locate reasoning errors in student solutions 
even with access to reference solutions 
\citep{srivatsa-etal-2025-llms-spot, 
tyen2024llmsreasoningerrorscorrect, estevez2025evaluation}.
Yet these limitations remain unaddressed in 
structured symbolic domains where step-level 
feedback alignment is especially consequential.

\subsection{Role-specialized Feedback Pipelines}
Role-specialized pipelines have shown promise in 
educational AI, where role separation ensures 
feedback meets both accuracy and pedagogical 
standards \citep{daheim2024stepwise, phung2024automating}, addressing over-praise and over-inference errors \citep{dos2025intelligent, chudziak2025ai}. Prior work shows that role-specialized multi-agent systems improve instructional design \citep{zhang2025eduplanner}, and that educational evaluation benefits from incorporating multiple pedagogical roles \citep{chen2025multi}. EducationQ further demonstrates that student–teacher–evaluator triads promote effective scaffolding and align with human pedagogical judgment \citep{shi-etal-2025-educationq}. Yet these limitations remain unaddressed in structured symbolic domains where step-level 
feedback alignment is especially consequential.

Beyond educational context, prior work shows that self-correction without external grounding can degrade reasoning performance \citep{huang2024selfcorrect}. Intrinsic self-correction sometimes worsens performance across diverse reasoning tasks, including arithmetic reasoning, code generation, and question answering \cite{kamoi2024can}, and that verification effect may help weak generators while offering limited or negative gains for strong ones in general reasoning tasks \citep{zhang2024small}. However, these findings are largely confined to general reasoning tasks, and whether these patterns hold in step-level tutoring contexts remains an open question.

\subsection{Logic Reasoning Benchmarks}

Existing benchmarks for symbolic or logical reasoning primarily evaluate end-to-end proof generation using single-reference solutions, with limited or no annotations for intermediate reasoning steps \citep{saparov2022language}. ProofWriter \citep{tafjord2021proofwriter} assesses only final proof validity, obscuring whether models apply inference rules correctly at each step. FOLIO \citep{han2024folio} provides expert-authored first-order logic problems but penalizes semantically valid alternative derivations, while ProntoQA \citep{saparov2022language} enables controlled evaluation through synthetic reasoning chains but cannot capture branching decision points where multiple inference rules may apply.

In contrast, step-level logic 
tutoring requires distinguishing incorrect rule use, 
suboptimal strategies, and state misalignment (steps 
inconsistent with the current proof state),
that existing benchmarks collapse into 
binary outcomes obscuring intermediate reasoning errors \citep{saparov2022language, 
tafjord2021proofwriter, chi2014icap}. These limitations motivate our KG-grounded framework 
(Section~\ref{sec:dataset}), which encodes not only 
\emph{whether} a step is derivable but \emph{how}; capturing the inference rule and parent statements justifying each derivation, enabling evaluation that existing single-reference benchmarks cannot support.

\begin{table}[t]
\centering
\setlength{\tabcolsep}{3pt}
\resizebox{\columnwidth}{!}{%
\begin{tabular}{lccccc}
\toprule
\textbf{Dataset} & \textbf{Task} & 
\textbf{Step} & \textbf{Branching} & 
\textbf{Student} & \textbf{Difficulty} \\
\midrule
ProofWriter~\citep{tafjord2021proofwriter} 
    & Proof Gen. & \xmark & \xmark & \xmark & \xmark \\
FOLIO~\citep{han2024folio} 
    & NLI & \xmark & \xmark & \xmark & \xmark \\
ProntoQA~\citep{saparov2022language} 
    & CoT & \cmark & \xmark & \xmark & \xmark \\
LogicLearner~\citep{inamdar2025logiclearner} 
    & Proof Gen. & \xmark & \xmark & \xmark & \xmark \\
\midrule
\textbf{Ours} 
    & Next-Step & \cmark & \cmark & \cmark & \cmark \\
\bottomrule
\end{tabular}}
\caption{\footnotesize Comparison of existing reasoning benchmarks.
\textit{Step}: step-level evaluation.
\textit{Branching}: annotated solution space at multiple branches.
\textit{Student}: real learner data.
\textit{Difficulty}: instance-level complexity metrics.}
\label{tab:dataset_comparison}
\end{table}

%% file: latex/3_dataset.tex
\section{Task and Dataset} \label{sec:dataset}
\subsection{Task Formulation} 

A propositional proof problem is a tuple $(\mathcal{P}, C)$, where
$\mathcal{P} = \{p_1, \ldots, p_n\}$ is a set of premises which are logical statements 
assumed to hold, and $C$ is
the target conclusion. A \textbf{proof state}
$\sigma = (\mathcal{P}, I, C)$ represents a student's progress, where
$I$ is the ordered set of intermediate statements 
derived so far by applying inference rules 
$r \in R$ (e.g., Modus Ponens, Disjunctive Syllogism; complete rule set in Appendix~\ref{appendix:rules}). Each $r$ derives a new statement 
from parent statements in $\mathcal{P} \cup I$, with $I = \emptyset$ initially.

Given $\sigma = (\mathcal{P}, I, C)$, the task is to predict the optimal
next step $s$,  a statement 
derivable in exactly one rule application from 
$\mathcal{P} \cup I$, that minimises remaining derivation distance to $C$, 
preferring shorter proofs that reflect more direct 
reasoning paths.
We focus on single-step prediction, as each feedback round targets 
exactly one student decision point. A representative proof instance is in
Appendix~\ref{appendix: example_state}.


\subsection{Dataset Construction} 

We draw on interaction logs from a propositional logic tutoring system used in an undergraduate discrete mathematics course, at a large public U.S.\ university. Further details and illustrations of the tutor are present in Appendix \ref{appendix: DT}. The tutoring system comprises seven levels: a diagnostic assessment level (pre-test), five practice levels, which include on-demand context-specific hints along with immediate feedback on correctness, and a summative assessment level (post-test). 


Our dataset comprises \textbf{516 unique proof states}, 
defined at the $(\mathcal{P}, I, C)$ triple level, from 32 problems across five practice levels of varying difficulty from Spring 2023. Each proof state instance represents a distinct student proof state, from minimal configurations with only three premises 
to complex states with up to 12 intermediate statements. Table \ref{tab:dataset} provides the complete distribution of proof states in the dataset, where unequal instance counts across levels reflect natural student activity. 


\subsection{Knowledge Graph Construction} \label{sec:KG}
We encode the complete solution space for each of 
the 32 problems as a knowledge graph $K$ capturing 
all valid inference paths from $\mathcal{P}$ to $C$. 
This encoding is exhaustive by construction: given 
a finite premise set and closed rule set $R$, all 
reachable derivations are enumerable via forward 
chaining \citep{barnes2008toward}.


Let $\Sigma$ be the set of all reachable proof states for each problem. The
knowledge graph $K = (\Sigma, E)$ has nodes $\sigma \in \Sigma$ and
directed edges $(\sigma_i, \sigma_j) \in E$ representing valid
single-step inferences. Each $\sigma_j$ is obtained from $\sigma_i$ by deriving a new intermediate statement $s$ via rule $r \in R$ applied to one or two
parent statements from $\mathcal{P} \cup I_i$, 
such that $I_j = I_i \cup \{d\}$. The graph is rooted at the initial state
$\sigma_0 = (\mathcal{P}, \emptyset, C)$, in which no intermediates have
yet been derived, and terminates at the state $\sigma^*$ in which $C$
has been successfully derived i.e., $C \in I^*$. Each edge is
annotated with the inference rule and the parent
statements used in the derivation.

\noindent\textbf{Next-step classification:}
For any state $\sigma \in \Sigma$, its distance from conclusion is
$d(\sigma) = \min_{\pi:\,\sigma \rightsquigarrow \sigma^*} |\pi|$,
computed via breadth-first search. A predicted step $\hat{s}$ from
state $\sigma_t$ yields successor $\hat{\sigma}_{t+1}$ and is
classified as:
\begin{enumerate}[topsep=0pt, partopsep=0pt, 
noitemsep, leftmargin=*]
\item \textbf{Optimal:} $(\sigma_t, \hat{\sigma}_{t+1}) 
\in E$ and $d(\hat{\sigma}_{t+1}) = d(\sigma_t) - 1$, 
i.e., a valid single-step inference that strictly 
reduces distance to the goal.
\item \textbf{Invalid:} $(\sigma_t, \hat{\sigma}_{t+1}) 
\notin E$, i.e., the predicted step is not a valid 
single-step inference from the current state, a subtle failure we illustrate below.
\end{enumerate}

\noindent A predicted step $\hat{s}$ 
may be symbolically valid, yet inferentially unjustified if it is attributed to an incorrect rule or unsupported parent statements. For example, a model predicting statement $C$ via Modus Ponens must correctly identify both $(A \rightarrow C)$ and $A$ as parents; a prediction that produces a valid statement but misattributes the rule or parents is flagged as unjustified regardless of symbolic validity. 

This distinction is pedagogically consequential: tutoring feedback that validates a correct step arrived at via flawed reasoning reinforces the misconception rather than correcting it. We call it partial correctness validation failure mode we identify in Section \ref{sec:rq3}.

%% file: latex/4_method.tex
\section{Role-Specialized Tutoring Pipelines}
\vspace{-0.1em}
\label{sec:method}
We distinguish three role-specialized pipelines 
that vary in solution information access: 
Student-Tutor, Student-Teacher, and 
Student-Tutor-Judge. In each pipeline, agents 
operate sequentially in a fixed order with no 
bidirectional communication as illustrated in Fig \ref{fig:framework}. 


We operationalize solution information access 
as a controlled variable across three agents:
\begin{itemize}[topsep=0pt, partopsep=0pt, leftmargin=*, nosep]
    \item \textbf{Tutor}: receives only correct next step $s$. 
    \item \textbf{Teacher}: receives correct next step $s$ with complete derivation context (inference rule, parents).
    \item \textbf{Judge}: receives complete derivation context as the Teacher along with Tutor feedback.
\end{itemize}
\noindent This asymmetric grounding isolates (i) \emph{information access} (Tutor vs.\ Teacher) and (ii) \emph{verification effects} (Tutor vs.\ Judge) as independent variables controlling for task difficulty. 

\subsection{Student}

The Student agent simulates an undergraduate learner with no solution 
access following \citep{shi-etal-2025-educationq}. Given a proof state $\sigma_t$ the agent predicts the next state $\sigma_{t+1}$. To this end, it generates (i) \textit{2--3 candidates with rationales}, (ii) \textit{select the optimal $\sigma_{t+1}$ along reasoning}, (iii) \textit{next step $s$}, (iv) \textit{inference rule}, and (v) \textit{parent statements}. This multi-candidate approach reflects least-to-most prompting \citep{zhou2022least} and pedagogical chain-of-thought \citep{jiang2024llms}, externalizing reasoning that tutoring agents can reference when scaffolding.

After feedback, the Student agent revises their response only if errors are identified ensuring measured improvements reflect genuine feedback effects rather than forced changes. Student agent prompts and a sample student response are shown in Figs   \ref{fig:student-prompt}, \ref{fig:student-update-prompt}, and \ref{fig:student-response}.

The Student agent revises only when errors are identified, ensuring measured improvements reflect genuine feedback effects rather than forced changes.

\subsection{Tutor}

The Tutor models a teaching assistant (TA) who knows the correct answer but not the derivation, a more constrained setting than the restricted-teacher condition in \citet{shi-etal-2025-educationq}. Tutor receives only the optimal next step $s$ without derivation context (rule, parent statements), requiring inference of both the applicable rule and reasoning \citep{kochmar2022automated}. It classifies student responses as optimal or invalid and provides 
scaffolded feedback guided by the 
shared scaffolding constraints (§4.5). This tests whether correctness-only grounding suffices for effective tutoring. Tutor prompt and a sample response are shown in Fig \ref{fig:tutor-prompt} and Fig \ref{fig: Tutor-response}. 

\subsection{Teacher }
The Teacher models an instructor with full solution access, the standard assumption in ITS \citep{corbett2001cognitive, barnes2008toward} representing the upper bound on information availability. It receives the complete 
derivation $\sigma_{t+1}$ (rule, parents, optimal next step $s$), enabling precise evaluation and feedback, which may risk answer revelation
\citep{daheim2024stepwise}, a limitation that we discuss in section \ref{sec:rq3}. Comparing with Tutor, Teacher isolates the effect of solution access on feedback quality. Teacher prompt and sample responses are shown in Fig \ref{fig:teacher-prompt} and Fig \ref{fig:teacher-response}.

\subsection{Judge}

The Judge models a quality-assurance reviewer, following verifier-in-the-loop designs in educational AI \citep{phung2024automating, shi-etal-2025-educationq}. It receives full derivation context $\sigma_{t+1}$ and Tutor feedback, to evaluate both the student's 
step and the Tutor's feedback for correctness and 
answer leakage. Based on this, it either enhances or overrides the tutor's feedback, addressing over-praise and answer-leakage failures common in single-agent systems \citep{gonnermann2025facet, zhang2025eduplanner}. Judge prompt and sample responses are given in Fig \ref{fig:judge-prompt}, Fig \ref{fig:tutor-ambiguous-verification}, and \ref{fig:tutor-echo-verification}.
\vspace{-0.1em}
\subsection{Shared Constraints}
\vspace{-0.1em}
All feedback agents adhere to scaffolding principles from learning science, guide discovery without revealing answers \citep{ belland2017scaffolding, vanpol2010scaffolding}, employ Socratic questioning over directives \citep{chen2023its_sustainable}, reference student candidates for relevance \citep{serban2020korbit}, and acknowledge correct reasoning before addressing errors \citep{shi-etal-2025-educationq}. Feedback is limited to 2-3 sentences. 

\subsection{Interaction Protocol}

Each interaction consists of a single feedback round to isolate pipeline effects. To ensure fair comparison, initial Student responses were generated once and reused across all pipelines; Tutor feedback from the Student-Tutor pipeline was reused in the Student-Tutor-Judge pipeline. All agents communicate via structured JSON format, where fewer than 3\% of instances required a retry due to malformed JSON output. Implementation details and quality validation constraints across dialogue generation appear in Appendix \ref{appendix:implementation}.

%% file: latex/5_evaluation_metrics.tex
\section{Experimental Setup and Evaluation}

\subsection{Evaluation Metrics}\label{sec:evaluation}

Building on the KG evaluation framework, we define six metrics that 
operationalize pipeline performance at the 
proof-state level.

\noindent\textbf{Task Performance:} We report 
\textit{Pre} (student-only) and \textit{Post} 
(after feedback) accuracy, each evaluated over 
all $N{=}516$ proof states. It is
computed as the proportion of Optimal predictions by classifying predicted next steps as 
Optimal or Invalid per next-step classification described in (Sec \ref{sec:KG}). The \textit{Learning gain} $\Delta = 
\text{Post} - \text{Pre}$ is reported in 
percentage points (pp).

\noindent\textbf{Tutor Rule Accuracy:} measures 
how often the Tutor correctly predicts the 
inference rule for the optimal step, given only 
$s$ as input, serving as a proxy for upstream 
feedback reliability.

\noindent\textbf{Reasoning Difficulty Analysis:}
\textit{Step Complexity} quantifies the symbolic 
difficulty of a proof step as a weighted function 
of logical operators and nesting depth 
(Appendix~\ref{appendix:KG_metric}), providing a 
model-independent characterization of instance 
difficulty. We report \textit{Mean Complexity}, 
the average Step Complexity of proof states where 
each pipeline improves student performance, to 
characterize which difficulty regimes benefit from 
each condition. 

\noindent\textbf{Verification Effects:}
To isolate verification's effects
beyond full solution access, we compute the 
\textit{Unique Improvement Count (UIC)}, the number of proof states where only the Judge pipeline succeeds and the Teacher pipeline fails on identical student inputs. Since both pipelines share full derivation access, any Judge improvement over Teacher is attributable solely to verification rather than information access.

\noindent\textbf{Feedback Quality (Human Annotation):}
To identify failure mechanisms that automated metrics cannot 
capture, three domain experts with at least two years of 
undergraduate Discrete Structures teaching experience 
independently annotated feedback responses across all 
pipelines on four dimensions: \textit{Mistake 
Identification}, \textit{Correctness}, \textit{Answer 
Revelation}, and \textit{Actionability} 
(Krippendorff's $\alpha \geq 0.65$; full protocol in 
Appendix~\ref{appendix:human-eval}). Three systematic 
failure modes emerge.

%% file: latex/6_experiment_setup.tex
\subsection{Experimental Setup}
\label{experiment}

We evaluate seven LLMs spanning reasoning-augmented models (GPT-o3, Gemini-1.5-Pro, DeepSeek-R1), standard instruction-tuned models (GPT-4.1, Qwen-3-32B), and open-source models 
(Mistral-Large, LLaMA-3.3-70B). This selection covers diverse capability levels and architectural families, enabling analysis of whether pipeline effects are 
capability-dependent (Sec \ref{discussion}).

Each model serves as both a student and a feedback agent (Tutor, Teacher, Judge) across all pipelines, generating $516 \times 7 \times 3 = 10{,}836$ 
dialogues (516 proof states $\times$ 7 models 
$\times$ 3 pipelines). All experiments use temperature 0.0 via APIs listed in Appendix \ref{appendix:model-specs} for reproducibility. Model specifications and sample dialogues are presented in Appendices \ref{appendix:model-specs} and \ref{appendix:agent-responses}.

%% file: latex/7_results.tex
\section{Results} \label{discussion}
We organize our findings around three RQs: how information 
access shapes feedback (RQ1), when does role-specialized verification improves or degrades the feedback (RQ2), what failure modes emerge in feedback across pipelines and whether automated metrics capture the pedagogical impact (RQ3). We begin with baseline performance, then address each RQ in turn, complimenting the evaluation metrics with expert human evaluation of feedback, and discussing the pedagogical aspects.

\begin{table*}[t]
\setlength{\tabcolsep}{4pt}
\centering
\footnotesize
\caption{Pipeline performance across seven LLMs ($N = 516$), bold values indicate best Post accuracy per model.}
\label{tab:main-results}
\begin{tabular}{lcccccccccccc}
\toprule
\textbf{Model} &
\textbf{Pre (\%)} &
\textbf{Rule Acc.} &
\multicolumn{3}{c}{\textbf{Post (\%)}} &
\textbf{$\Delta$} &
\multicolumn{3}{c}{\textbf{Mean Complexity}} &
\multicolumn{2}{c}{\textbf{UIC over Teacher}} \\
\cmidrule(lr){4-6}
\cmidrule(lr){8-10}
\cmidrule(lr){11-12}
& 
\textbf{Student} &
\textbf{Tutor (\%)} &
\textbf{Tutor} &
\textbf{Teacher} &
\textbf{Judge} &
\textbf{pp} &
\textbf{Tutor} &
\textbf{Teacher} &
\textbf{Judge} &
\textbf{UIC} &
\textbf{Gap} \\
\midrule

\midrule
\textbf{GPT-4.1} &
21.31 & 58.30 &
42.82 & 61.24 & \textbf{75.38} &
54.07 &
1.8 & 2.21 & 2.7& 89 & 1.63\\

\textbf{GPT-o3} &
24.80 & 60.10 &
48.25 & 61.24 & \textbf{75.58} &
50.78 &
2.04 & 2.26 & 2.65 & 84 & 1.4\\

\textbf{Gemini-1.5-pro} &
52.52 & 85.8 &
73.92 & \textbf{78.79} & 72.95 &
20.43 &
3.73 & 3.17 & 3.65 & 23 & 2.22\\

\textbf{Deepseek-R1} & 56 & 84.9 & 76.55 & \textbf{84.30} & 79.84 & 23.84 & 3.49 & 3.06 & 3.37 & 17 & 3.19\\

\textbf{Mistral-Large} &
23.74 & 54.80 &
38.41 & 41.69 & \textbf{56.37} &
32.63 &
1.86 & 1.85 & 2.43 & 103 & 0.95\\

\textbf{Qwen-3-32B} & 46 & 70.20 & 60.36 & 63.63 & \textbf{66.90} & 20.9 & 2.29 & 2.19 & 2.6 & 39 & 1.59\\
\textbf{Llama-3.3-70B} & 17.98 & 24.70 & 24.30 & 34.38 & \textbf{39.32} & 21.34 & 2.06 & 2.06 & 2.33 & 55 & 0.41\\
\bottomrule
\end{tabular}
\footnotesize
\RaggedRight
\textbf{Pre}/\textbf{Post}: student accuracy before and after feedback;
$\Delta$: learning gain in percentage points (pp);
\textbf{Rule Acc.}: Tutor rule prediction accuracy, proxy for upstream reliability;
\textbf{Mean Complexity}: average proof-state complexity of pipeline-improved instances;
\textbf{UIC}: instances where Judge succeeds after Teacher fails;
\textbf{Gap}: difference between predicted and optimal step complexity (0 is ideal).
\end{table*}

\subsection{Baseline (Student Reasoning Performance)}
As shown in Table~\ref{tab:main-results}, baseline performance varies markedly across models. DeepSeek and Gemini achieve the highest pre-feedback accuracy (56\% and 54.5\%), followed by Qwen (46\%), while GPT variants and Mistral perform more modestly (20–25\%) and LLaMA performs weakest (17\%). Examining baseline accuracy alongside Tutor rule accuracy reveals two model groups. We refer to Gemini and DeepSeek, which consistently exhibit high performance on both measures, as \emph{strong models}. In contrast, GPT models, Mistral, Qwen, LLaMA show lower performance on both dimensions and are referred to as \emph{weak models} in this paper.

\subsection{Effect of Information Access on Feedback 
Quality (RQ1)}
\label{sec:rq1}
\paragraph{Tutor:}
Tutor agents receive only the correct next step without rule 
or parent statements, requiring inference of the full 
derivation before generating feedback. Despite this, Tutor post-feedback accuracy 
reaches $24-77\%$ across models, with models referred to as weak improved 
modestly over baseline (GPT-4.1: 21\%~$\to$~43\%, 
Mistral: 24\%~$\to$~38\%, LLaMA: 18\%~$\to$~24\%) and 
strong models improving more substantially 
(Gemini: 53\%~$\to$~74\%, DeepSeek: 56\%~$\to$~77\%). 
However, these accuracy gains mask a systematic limitation: 
without derivational context, Tutor feedback is often 
directionally correct but insufficiently specific to repair 
student reasoning at the rule level. Human evaluation further confirms this limitation, as discussed in Sec \ref{sec:rq3}.

\noindent\textbf{Teacher:}
Full derivation access (Teacher) improves feedback on harder 
instances for some models, but not uniformly. Among improved 
cases, Teacher mean complexity (1.85-3.17; 
Table~\ref{tab:main-results}) exceeds Tutor primarily for 
GPT-based models, while Tutor reaches comparable or higher 
complexity for others. Logistic regression confirms that 
Teacher $\times$ Complexity interactions are significant 
mainly for strong models (Gemini: $\beta=+0.65$, 
DeepSeek: $\beta=+0.36$, $p<.001$; 
Table~\ref{tab:logit-effects}), indicating that full access 
benefits only models able to utilize derivational context. 



\subsection{Conditional Effects of Verification (RQ2)}
\label{sec:rq2}

Table~\ref{tab:main-results} shows that verification yields 
large gains for weak Tutor agents (Rule Acc.\ $<$70\%: 
GPT-4.1, GPT-o3, Mistral, LLaMA), averaging +35pp, but 
\emph{degrades} performance for strong Tutor agents 
(Gemini, DeepSeek; Rule Acc.\ $>$84\%) by approximately 
$-5pp$ as compare to the Teacher. This asymmetry correlates with complexity: weak 
Tutor agents improve only on low-complexity problems 
(mean 1.8--2.1), while Judge agents extend gains to harder 
problems (mean 2.4--2.7). Strong Tutor agents already 
succeed at high complexity (mean 3.5--3.7), leaving limited 
scope for verification.

\textbf{Unique Improvement Count}
The UIC quantifies verification's marginal contribution 
beyond Teacher: weak models show $n \in [84, 103]$ unique Judge 
successes after Teacher failure, vs.\ 17--23 for strong 
models (Table~\ref{tab:main-results}). The Complexity Gap 
(0.4--1.6 for weak, 2.2--3.2 for strong models) confirms 
that verification addresses high-complexity edge cases where 
weak Tutor feedback fails. Regression analysis confirms 
this pattern: Judge $\times$ Complexity interactions are 
positive and significant across nearly all models 
(GPT-4.1: $\beta=+0.22$, Gemini: $\beta=+0.61$, 
DeepSeek: $\beta=+0.56$; all $p<.001$; 
Table~\ref{tab:logit-effects}).

\textbf{Ablation: Verifier on Teacher Feedback.}
To further isolate the role of verification, we applied 
the same Judge agent to Teacher feedback to test whether 
verification can mitigate premature revelation. This yields 
no consistent benefit: performance improves only for 
DeepSeek (+4pp) and is neutral or harmful for all other 
models. 

These results confirm that verification 
effectiveness depends on upstream \emph{error rate}, not 
\emph{information access}: the Judge corrects erroneous 
guidance but cannot enhance already reliable feedback, and 
cannot undo premature revelation once disclosed. Human 
evaluation is consistent with this: Judge feedback scores 
higher on \textit{Correctness} and \textit{Actionability} 
than Tutor for weak models. \textcolor{red}{mention the radar plot}

\textbf{Universal Complexity Ceiling}
Across all models and pipelines, performance degrades 
sharply once proof-state complexity exceeds 4--5 
(Table~\ref{tab:degrade} in Appendix). This ceiling is 
independent of verification or information access: simple 
states (complexity $<$2.5) require minimal intervention, 
mid-complexity states (2.5--4.0) benefit most from 
verification, and high-complexity states ($>$4.5) remain 
challenging for all approaches, suggesting that the 
bottleneck at high complexity is symbolic reasoning 
capacity rather than feedback pipeline design.

\subsection{Human Evaluation - Failure Mode Analysis (RQ3)}
\label{sec:rq3}
Automated accuracy metrics reward correct outcomes 
but cannot penalize pedagogically flawed reasoning 
paths. Human evaluation across four dimensions 
(§\ref{sec:evaluation}) reveals four systematic 
failure modes.
\textbf{Echo Effect (Tutor).}
Weak Tutor agents, lacking derivational context, 
anchor on the student's incorrect rule rather than 
correcting it (Fig~\ref{fig:tutor-echo-verification}). 
Human evaluation confirms this systematically: Tutor 
feedback scores lowest on \textit{Mistake 
Identification} among weak models, indicating 
insufficient derivational grounding rather than 
random noise. Verification mitigates this through 
KG-grounded override (§\ref{sec:rq2}).

\textbf{Partial Correctness Validation (Tutor).}
When students predict the correct step via flawed 
reasoning, Tutor agents validate the faulty logic, 
reinforcing misconceptions. \textit{Correctness} 
scores remain moderate even when post-feedback 
accuracy improves, confirming that accuracy 
overestimates genuine reasoning repair 
\citep{kamoi2024can, tyen2024llmsreasoningerrorscorrect}.

\textbf{Premature Revelation.}
Teacher feedback frequently discloses the correct step or 
over-specifies derivational hints 
(Fig~\ref{fig:teacher-reveals-response}), leading student 
agents to anchor on the disclosed answer without 
integrating it into their reasoning. Human evaluation 
confirms that Teacher feedback scores highest on 
\textit{Answer Revelation} while \textit{Actionability} 
scores remain moderate, indicating that disclosed answers 
produce surface-level correctness rather than transferable 
reasoning improvement.

\textbf{Verification Over-Specification.}
Although Judge feedback improves \textit{Correctness} and 
\textit{Mistake Identification} for weak models, human 
evaluation reveals a distinct failure mode for strong models: 
when the Judge overrides already reliable Tutor feedback, the 
resulting guidance becomes overly abstract, reducing 
\textit{Actionability} scores below those of the original 
Tutor feedback. This pattern mirrors the quantitative 
degradation observed for Gemini and DeepSeek ($-$5pp) and 
suggests that verification introduces a specificity cost 
when upstream feedback is already well-grounded: the Judge 
imposes derivational constraints that narrow rather than 
scaffold the student's reasoning space.

Human evaluation confirms the four failure modes, demonstrating that automated metrics overestimate feedback quality by rewarding correct outcomes without penalizing flawed reasoning paths 
or answer leakage.

%% file: latex/8_conclusions.tex
\section{Conclusion}
\label{sec:conclusion}
We present a controlled study of multi-agent LLM tutoring for propositional logic using a knowledge-graph–grounded benchmark of 516 proof states. Our central finding is that verification benefit is \emph{asymmetric}: it improves outcomes by +10–14pp when upstream feedback is error-prone (<70\% accuracy), but degrades performance by 4–6pp when feedback is already reliable (>85\%). This asymmetry is governed by the proof-state Complexity. Verification helps in intermediate regimes but cannot overcome a shared reasoning ceiling (complexity >4–5) beyond which no pipeline reliably succeeds. We identify three systematic failure modes: echo effects, partial correctness validation, and premature revelation that verification mitigates only under limited conditions. These findings challenge the assumption that adding verifiers or solution access improves tutoring. Instead, effective LLM tutoring requires adaptive pipelines that estimate upstream reliability and route problems by complexity, rather than defaulting to more agents or richer context.

%% file: latex/limit_future.tex
\section*{Limitations}

This study has several limitations. First, our evaluation is restricted to propositional logic proof tutoring, and the findings may not directly generalize to domains with weaker symbolic constraints or richer semantic structure. While our primary metrics assess the intrinsic properties of feedback independently of the receiver, LLM students may respond differently from human learners; therefore, human-in-the-loop validation would be necessary to substantiate claims about learning outcomes. Second, while our knowledge-graph-based metrics enable fine-grained analysis of step-level reasoning, they rely on a predefined solution space and do not capture open-ended or creative reasoning strategies. Third, our experiments focus on single-step feedback and do not model long-horizon student–tutor interactions or learning gains over time. Moreover, our qualitative analysis is limited to a small expert-reviewed sample (10 responses per model), and broader manual analysis could reveal additional failure modes. Finally, although we analyze multiple large language models, results may vary with future architectures or training regimes.

\section*{Ethics Statement}
Student interaction data was collected under IRB approval with informed 
consent. All data was anonymized before analysis. Our LLM-based 
Evaluation does not involve human subjects beyond annotators, who serve as researchers for the tutoring system used for data generation in this study, as well as co-authors of this study. 

\section*{Reproducibility}

Code and data will be released upon publication. All experiments use temperature 0.0 to ensure deterministic, reproducible results, prioritizing controlled comparison over response diversity. While real-world tutoring systems might benefit from non-zero temperature, our evaluation focuses on measuring the systematic effects of multi-agent architectures under controlled conditions. Model specifications and 
prompts appear in Appendices \ref{appendix:model-specs} and \ref{appendix:prompts}.

\section*{AI Usage Disclosure}
This work studies and evaluates large language models as research objects. We utilized large language models as assistive tools during manuscript preparation, including formatting guidelines and brainstorming organization, mainly, and paraphrasing on a need basis only. We did not use any AI tools for designing, implementing, or executing this research study. All the claims, analyses, hypotheses, and conclusions are developed, verified, and reviewed by the authors. Moreover, no AI tool was used for generating or labeling data, making judgments about data, or making any scientific claims. All implementations were reviewed, validated, and finalized by the authors. The authors take full responsibility for the correctness, originality, and integrity of the work.

%% file: latex/appendix.tex
\appendix
\nolinenumbers

\section{Inference Rule List}
\label{appendix:rules}

We employ a fixed set of propositional inference rules used in Logic Tutor for the dataset. The short names were used for consistent response generation and evaluation. The complete list of inference rules, along with short names and derivations are provided below in Table \ref{tab:inference-rules}.

\begin{table}[H]
\centering
\small
\footnotesize
\caption{Propositional inference rules used in this work.}
\label{tab:inference-rules}
\setlength{\tabcolsep}{2pt}
\begin{tabular}{lll}
\toprule
\textbf{Abbrev.} & \textbf{Rule Name} & \textbf{Form} \\
\midrule
MP   & Modus Ponens 
     & $P \rightarrow Q,\; P \Rightarrow Q$ \\

MT   & Modus Tollens 
     & $P \rightarrow Q,\; \neg Q \Rightarrow \neg P$ \\

Conj & Conjunction 
     & $P,\; Q \Rightarrow P \land Q$ \\

Simp & Simplification 
     & $P \land Q \Rightarrow P$ (or $Q$) \\

Add  & Addition 
     & $P \Rightarrow P \lor Q$ \\

DS   & Disjunctive Syllogism 
     & $P \lor Q,\; \neg P \Rightarrow Q$ \\

HS   & Hypothetical Syllogism 
     & $P \rightarrow Q,\; Q \rightarrow R$ \\
     & 
     & $\Rightarrow P \rightarrow R$ \\

Impl & Implication 
     & $P \rightarrow Q \equiv \neg P \lor Q$ \\

DN   & Double Negation 
     & $P \equiv \neg\neg P$ \\

CP   & Contraposition 
     & $P \rightarrow Q \equiv \neg Q \rightarrow \neg P$ \\

Com  & Commutation 
     & $P \lor Q \equiv Q \lor P$ \\

Assoc& Associativity 
     & $(P \lor Q)\lor R$ \\
     & 
     & $\equiv P \lor (Q \lor R)$ \\

Dist & Distribution 
     & $P \land (Q \lor R)$ \\
     & 
     & $\equiv (P \land Q) \lor (P \land R)$ \\

CD   & Constructive Dilemma 
     & $(P\!\rightarrow\!Q),(R\!\rightarrow\!S),P\lor R$ \\
     & 
     & $\Rightarrow Q \lor S$ \\

Equiv& Equivalence 
     & $P \leftrightarrow Q$ \\
     & 
     & $\equiv (P\!\rightarrow\!Q)\land(Q\!\rightarrow\!P)$ \\

\bottomrule
\end{tabular}
\end{table}

\section{Representative Proof State}
\label{appendix: example_state}

This appendix presents an illustrative propositional logic proof instance Fig \ref{fig:proof-instance} annotated with intermediate derivations and agent-specific hints. Intermediate steps are derived by applying valid inference rules to previously established statements, with each step indexed and linked to its parent statements and rule application. This explicit structure exposes the shortest derivation path toward the conclusion while preserving alternative valid reasoning trajectories.

The Tutor is provided only with the KG-verified optimal next symbolic step, enabling guidance that focuses on local progression without access to derivational context. In contrast, the Teacher additionally receives the rule and parent statements used to derive this step, allowing feedback to reference structural reasoning without revealing the answer directly. This distinction isolates the effect of solution access: the Tutor operates under minimal grounding, while the Teacher leverages full derivational context to support more informed intervention. The representative proof state is available in Fig \ref{fig:proof-instance}.

\begin{figure}[t]
\centering
\begin{tcolorbox}[
    colback=white,
    colframe=black,
    boxrule=0.6pt,
    arc=2pt,
    left=6pt,
    right=6pt,
    top=6pt,
    bottom=6pt,
    title=\textbf{Representative proof Instance},
    fonttitle=\bfseries

]
\textbf{Givens}
\begin{itemize}[leftmargin=*, itemsep=0pt, topsep=2pt]
    \item (1) $((S \rightarrow Y) \lor (I*Q))$
    \item (2) $((I \land Q) \rightarrow D)$
    \item (3) $\neg D$
    \item (4) $((S \rightarrow Y) \rightarrow D)$
\end{itemize}
\textbf{Intermediate Steps}
\begin{itemize}[leftmargin=*, itemsep=0pt, topsep=2pt]
    \item (5) $\neg(S \rightarrow Y)$ \hfill [MT: (3), (4)]
    \item (6) $\neg(I \land Q)$ \hfill [MT: (3), (2)]
    \item (7) $(S \rightarrow Y)$ \hfill [DS: (6), (1)]
\end{itemize}

\textbf{Conclusion:} $Y$

\tcbsubtitle{\textbf{Tutor Hint}}
$\neg(\neg S \lor Y)$

\tcbsubtitle{\textbf{Teacher \& Judge Hint}}
Derive $\neg (\neg S \lor Y)$ from $\neg (S \rightarrow Y)$ using the Implication rule.
\end{tcolorbox}
\caption{Illustrative proof instance showing intermediate derivations and differential hint access for Tutor and Teacher agents.}
\label{fig:proof-instance}
\end{figure}

\section{Illustrative Logic Tutor Proof Interaction}
\label{appendix: DT}

\begin{figure}[h]
    \centering
    \includegraphics[width=\linewidth]{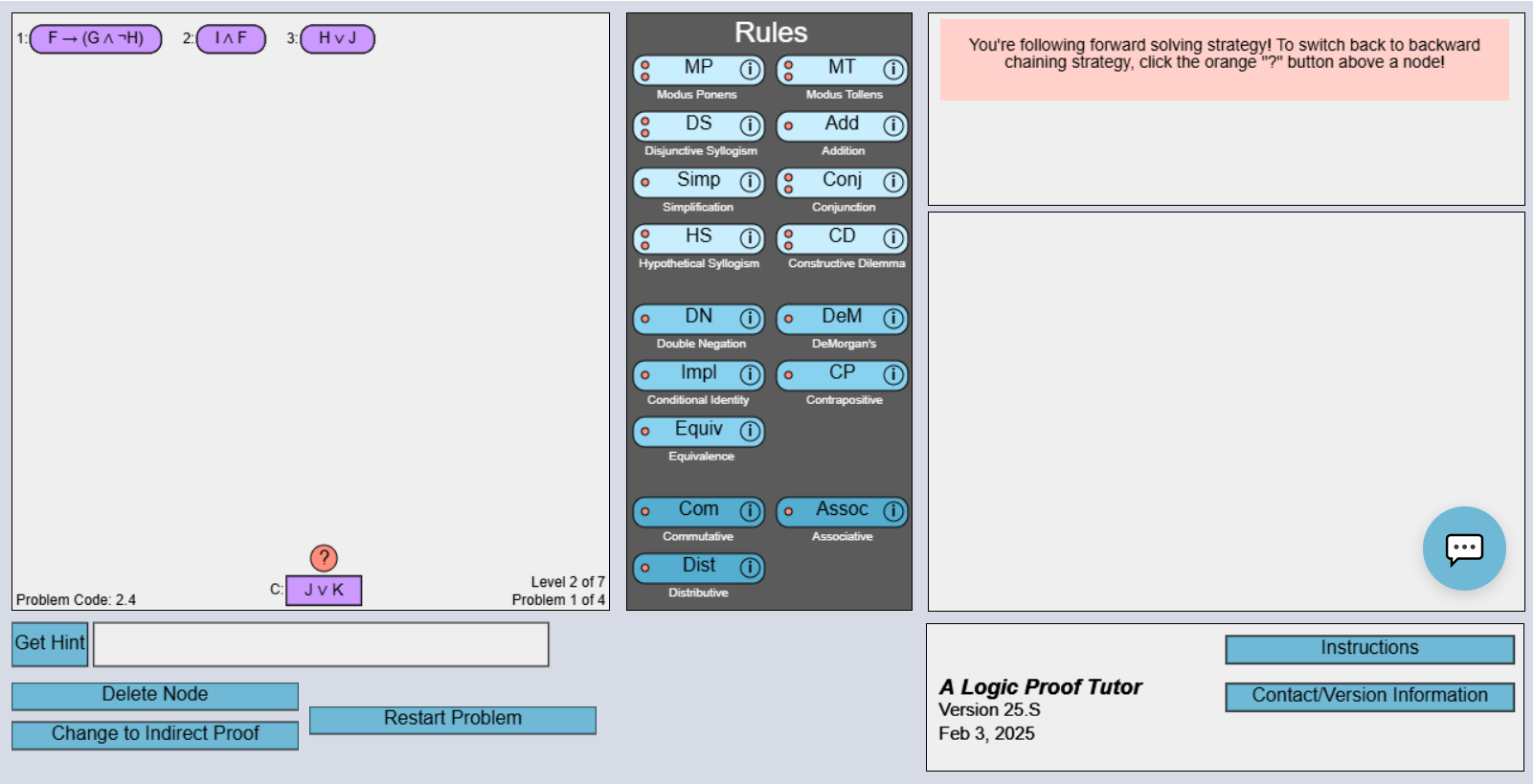}
    \caption{
    \textbf{Initial proof state and goal specification.}
    The student is presented with the premises (top left) and the target conclusion ($J \lor K$) at the bottom. Available inference rules are displayed on the middle. At this stage, no intermediate steps have been derived, and the student must choose a productive forward step toward the goal.
    }
    \label{fig:dt-initial}
\end{figure}

\begin{figure}[h]
    \centering
    \includegraphics[width=\linewidth]{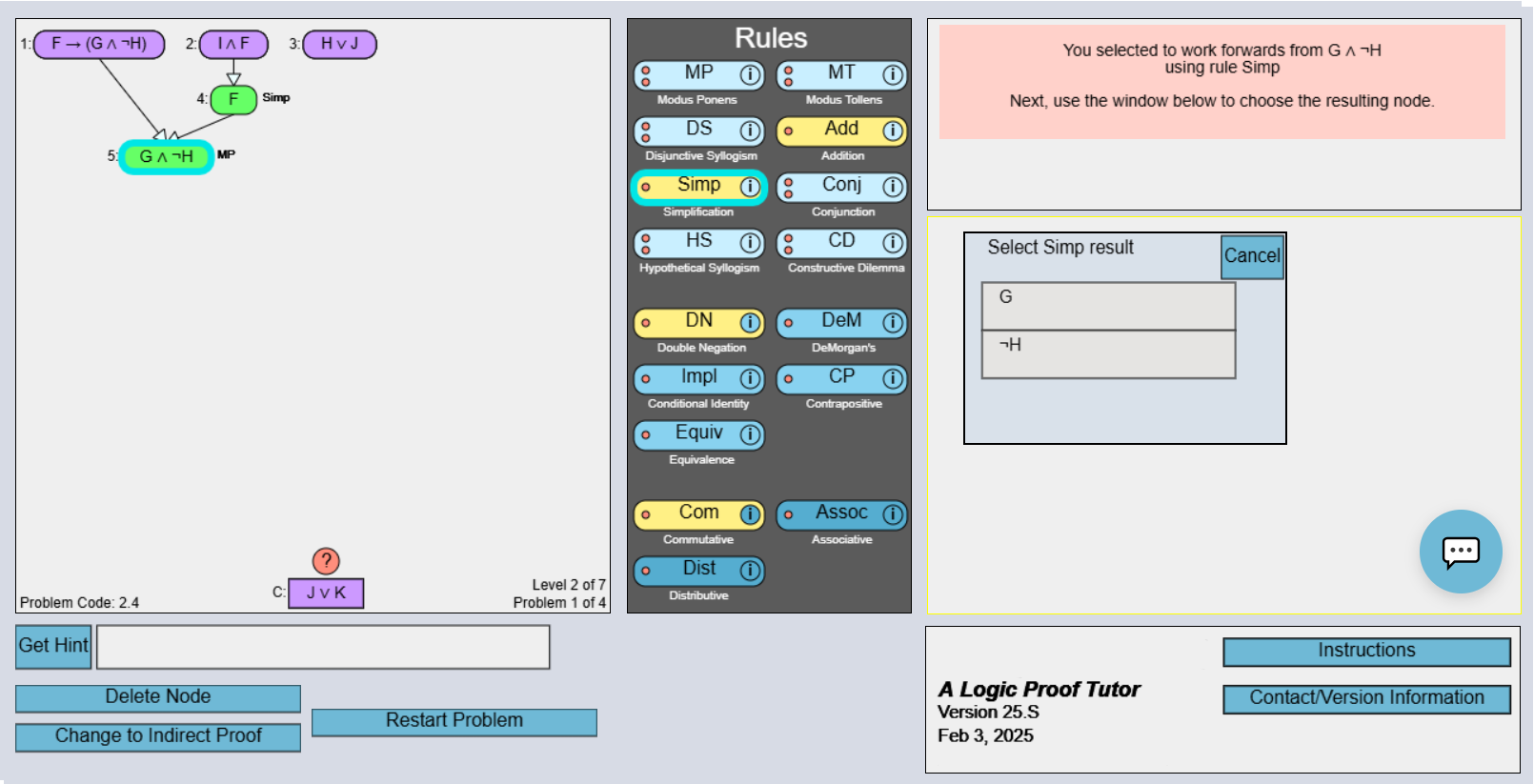}
    \caption{
    \textbf{Rule application with guided simplification.}
    After deriving an intermediate conjunction via Modus Ponens, the student applies the \textit{Simplification} rule. The interface prompts the learner to select the appropriate resulting literal ($G$ or $\lnot H$), illustrating fine-grained, step-level decision making supported by rule constraints.
    }
    \label{fig:dt-simplification}
\end{figure}

\begin{figure}[h]
    \centering
    \includegraphics[width=\linewidth]{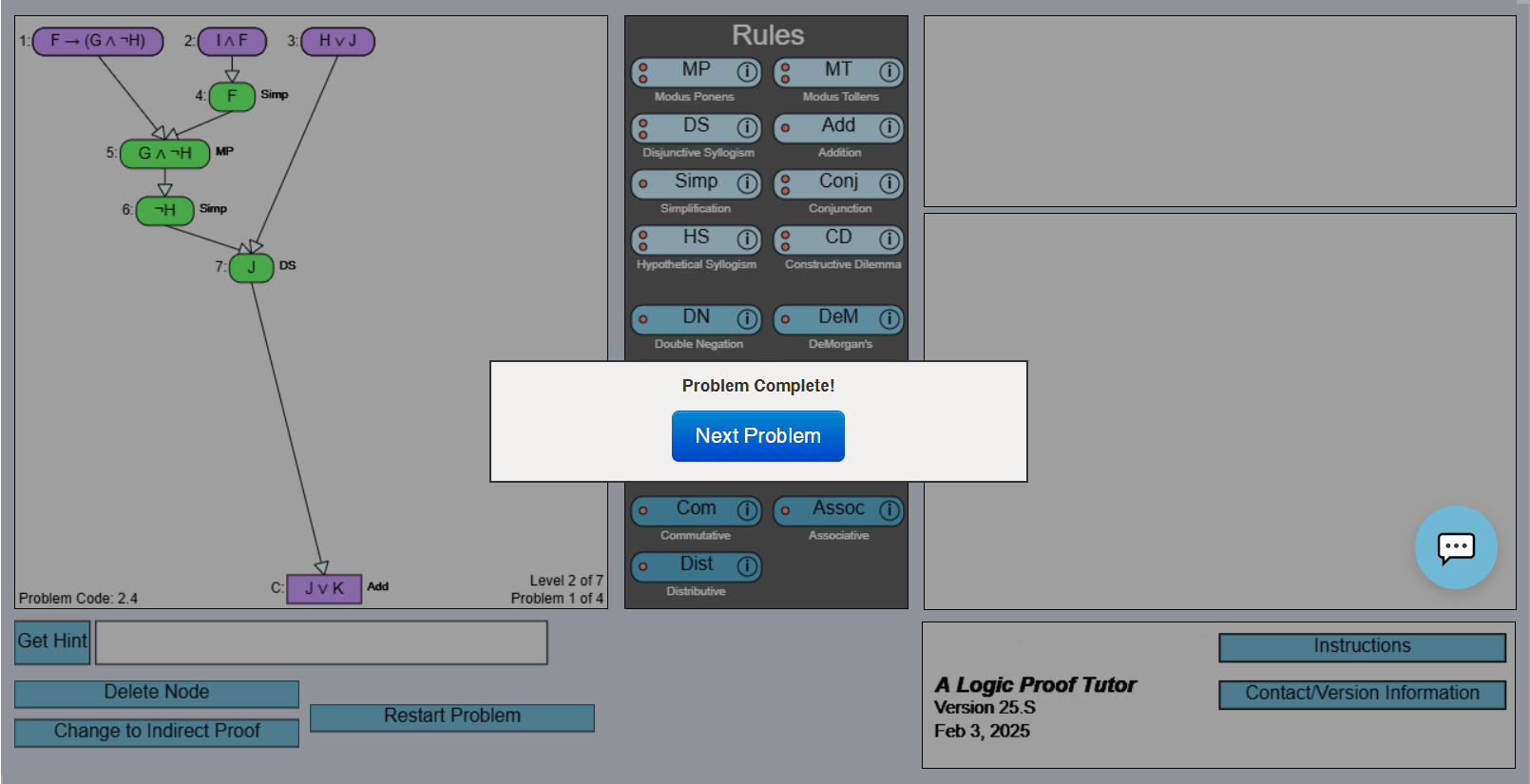}
    \caption{
    \textbf{Successful proof completion.}
    The student derives $J$ via Disjunctive Syllogism and applies the \textit{Addition} rule to reach the target conclusion $J \lor K$. The system confirms completion, reinforcing correct rule sequencing and alignment with the goal state.
    }
    \label{fig:dt-complete}
\end{figure}

Figure~\ref{fig:dt-initial}--\ref{fig:dt-complete} illustrates a representative student interaction in the propositional logic tutor. These screenshots demonstrate forward chaining, rule application, and goal completion within the tutor interface.

\section{Dataset Distribution}
\begin{table}[H]
\centering
\small
\caption{Dataset distribution across practice levels (2-6)}
\label{tab:dataset}
\begin{tabular}{lcc}
\toprule
\textbf{Difficulty} & \textbf{Proof States} & \textbf{Avg.\ Statements} \\
\midrule
2 (Introductory) & 48 & 5.86 \\
3 (Basic) & 111 & 7.73 \\
4 (Intermediate) & 148 & 7.94 \\
5 (Advanced) & 85 & 5.64 \\
6 (Expert) & 95 & 6.41 \\
\midrule
\textbf{Total} & \textbf{516} & \textbf{6.72} \\
\bottomrule
\end{tabular}
\vspace{+1mm}
{\footnotesize \textsuperscript{*}Levels 1 and 7 excluded because of unavailability of hints.}
\vspace{-4mm}
\end{table}

\section{Knowledge Graph based Evaluation Metrics} \label{appendix:KG_metric}

This appendix provides formal definitions and illustrative examples of the graph-based metrics used to evaluate next-step reasoning quality. All metrics are computed with respect to the knowledge graph (KG) constructed for each problem.

\subsection{Step Complexity}

Beyond derivational validity, \emph{Step Complexity} quantifies the intrinsic symbolic difficulty of a single inference step based on the structural complexity of the derived statement. We define step complexity as a function
\[
c : E \rightarrow \mathbb{R}^{+},
\]
computed directly from the syntactic structure of $\phi$. Let $\mathcal{O}(\phi)$ denote the set of logical operators in $\phi$, and let $\text{depth}(o)$ denote the nesting depth at which operator $o$ occurs. The complexity of the step is defined as:
\[
c(s_i, s_j) = \sum_{o \in \mathcal{O}(\phi)} w(o) \cdot \alpha^{\text{depth}(o)},
\]
where $w(o)$ is an operator-specific base weight (e.g., $\neg, \land, \lor, \rightarrow, \leftrightarrow$) and $\alpha > 1$ controls the penalty for nested structure. For example, $c(F) < c(A+B) < c(A \rightarrow (B+C)$, because no operator is involved, only 1 operator is involved, and 2 operators involved along with nesting.  

\subsection{Distance to Conclusion}
For any proof state $s \in \mathcal{S}$, its distance is defined as the length of the shortest derivation path to a terminal state $s^*$:
$d(s) = \min_{\pi : s \rightsquigarrow s^*} |\pi|$ computed via breadth-first search over the KG. A predicted next step is considered \emph{optimal} if it strictly reduces the distance to the goal by one.

\textbf{Example}  
If $d(s_t) = 4$ and a predicted successor state $\hat{s}_{t+1}$ satisfies $d(\hat{s}_{t+1}) = 3$, the step is distance-optimal. A valid step leading to a state with $d(\hat{s}_{t+1}) = 5$ is considered a valid it is non-optimal.

\subsection{Derivational Depth}
Derivational depth measures how many missing intermediate inference steps separate a predicted statement from the current proof state. This metric distinguishes immediately derivable steps from those that implicitly skip required reasoning and thus cannot be derived in one sentence. Formally, if for a predicted statement $\phi$ not present in the immediate successor states of $s_t$, we define derivational depth as the minimum number of edges required to derive $\phi$ from $s_t$ in the KG.

\textbf{Example}  
If $\phi$ can be derived from $s_t$ only after two intermediate inferences, its derivational depth is 2. Predictions with depth 1 correspond to directly derivable steps, while higher depths indicate premature or unsupported jumps in reasoning.

\subsection{Parent Statements and Rule Justification}
Each edge $(s_i, s_j) \in E$ in the KG is annotated with the inference rule applied and the minimal set of parent statements used in the derivation. This enables explicit verification of whether a model-generated step is justified by the available premises.

\textbf{Example}  
A model predicting statement $C$ with rule Modus Ponens must correctly identify both $(A \rightarrow C)$ and $A$ as parent statements. Predictions that produce a valid statement but fail to align with the correct parents or rule are flagged as unjustified.

\section{Prompt Design}
\label{appendix:prompts}
This section provides the complete prompts used for each agent in our multi-agent tutoring system. All prompts use structured JSON response formats to ensure consistent parsing.

Our prompt design translates established pedagogical principles into operational constraints for LLM agents. Each agent's prompt encodes specific instructional strategies grounded in learning science research, ensuring that generated feedback aligns with effective tutoring practices. We describe the key design decisions below.

\subsection{Student Prompt}
\label{appendix:student}

The base system message includes the Role and Task information along with instructions, whereas the user message includes the problem instance itself. The prompt template is provided below in Fig \ref{fig:student-prompt}.

\begin{figure}[H]
\centering
\begin{tcolorbox}[
    colback=white,
    colframe=black,
    boxrule=0.5pt,
    arc=2pt,
    left=3pt,
    right=3pt,
    top=3pt,
    bottom=3pt,
    title=\textbf{Student Prompt},
    fonttitle=\bfseries,
]
\textbf{Role:} You are a Student in an undergraduate Discrete Structures course
solving a propositional logic proof. Your task is to produce the \emph{single
most optimal next step} that advances the proof toward the conclusion.

\medskip
\textbf{Task:}
\begin{enumerate}[leftmargin=*, itemsep=0pt, topsep=0pt]
    \item Review the givens and intermediate steps.
    \item Propose 2--3 candidate next steps.
    \item Select the candidate that most directly advances toward the conclusion.
    \item Justify your choice and output the selected next step.
\end{enumerate}
\textbf{Constraints:}
\begin{itemize}[leftmargin=*, itemsep=0pt, topsep=0pt]
    \item Output exactly one next step in \emph{symbolic notation only}.
    \item Use only predefined inference rules (e.g., MP, MT, Conj, DS).
    \item Parent statements must be actual expressions, not line numbers.
\end{itemize}
\textbf{Response Format:}
\begin{itemize}[leftmargin=*, itemsep=0pt, topsep=0pt]
    \item \texttt{CANDIDATES}: 2--3 candidate steps with brief justification
    \item \texttt{REASONING}: Why the selected step is optimal
    \item \texttt{NEXT\_STEP}: Symbolic expression
    \item \texttt{RULE}: Inference rule (short name)
    \item \texttt{PARENT\_STATEMENTS}: Supporting expressions
\end{itemize}
\end{tcolorbox}
\caption{Base System Prompt for Student}
\label{fig:student-prompt}
\end{figure}

\subsection{Student Update Prompt}
\label{appendix:student-update}
After receiving feedback, the Student Update agent simulates a learner revising their approach. The prompt requires the student to: (1) identify what aspect of the feedback addresses their error, (2) generate revised candidates incorporating the guidance, and (3) produce a corrected next step.

Critically, if the original response was marked ``Correct,'' the agent responds with ``No Improvements Needed'' rather than artificially generating changes. This prevents the evaluation artifact of forced revisions when none are warranted, ensuring that measured improvement rates reflect genuine learning effects. The base system prompt is provided in Fig \ref{fig:student-update-prompt}.
\begin{figure}[H]
\centering
\begin{tcolorbox}[
    colback=white,
    colframe=black,
    boxrule=0.5pt,
    arc=2pt,
    left=3pt,
    right=3pt,
    top=3pt,
    bottom=3pt,
    title=\textbf{Student Update Prompt},
    fonttitle=\bfseries,
]
\textbf{Role:} You are a Student revising your previous next-step attempt
after receiving feedback on a propositional logic proof.

\medskip
\textbf{Task:}
\begin{enumerate}[leftmargin=*, itemsep=0pt, topsep=0pt]
    \item Review the feedback to determine whether errors are present.
    \item If the prior response is correct, return \texttt{No Improvement Needed}.
    \item Otherwise, revise your reasoning and propose exactly one improved next step.
\end{enumerate}
\textbf{Constraints:}
\begin{itemize}[leftmargin=*, itemsep=2pt]
    \item Revise only the \emph{immediate next step}; do not extend the proof.
    \item Output symbolic expressions only; use predefined inference rules.
    \item Do not infer or reveal the full solution.
\end{itemize}
\textbf{Response Format:}
\begin{itemize}[leftmargin=*, itemsep=0pt, topsep=0pt]
    \item \texttt{REVISED\_REASONING}: Brief reflection or \texttt{No Improvement Needed}
    \item \texttt{IMPROVED\_STEP}: Symbolic expression or \texttt{No Improvement Needed}
    \item \texttt{BETTER\_RULE}: Inference rule (short name)
    \item \texttt{PARENT\_STATEMENTS}: Supporting expressions or \texttt{No Improvement Needed}
\end{itemize}
\end{tcolorbox}
\caption{Base system prompt for student update}
\label{fig:student-update-prompt}
\end{figure}

\subsection{Tutor Prompt}
\label{appendix:teacher}
The Tutor operates under \textit{minimal hint access}: it receives only the correct next step without knowledge of the complete solution path. The prompt enforces a critical \textbf{planning-before-feedback} requirement: the Tutor must first generate an internal derivation plan explaining how the correct step is derived (identifying the rule and parent statements) before producing student-facing feedback. This design decision ensures the Tutor develops a genuine understanding of the solution rather than pattern-matching, aligning with research showing that scaffolded feedback improves instructional quality \citep{serban2020korbit}. The prompt is illustrated in Fig \ref{fig:tutor-prompt}.
\begin{figure}[H]
\centering
\begin{tcolorbox}[
    colback=white,
    colframe=black,
    boxrule=0.5pt,
    arc=2pt,
    left=3pt,
    right=3pt,
    top=3pt,
    bottom=3pt,
    title=\textbf{Tutor Prompt},
    fonttitle=\bfseries,
]
\textbf{Role:} You are a Tutor evaluating a student’s proposed next step
in a propositional logic proof, with access to the KG-derived optimal step.
\textbf{Task:}
\begin{enumerate}[leftmargin=*, itemsep=0pt, topsep=0pt]
    \item Analyze how the optimal step is derived (rule and parent statements).
    \item Evaluate the student’s candidates, reasoning, and chosen next step.
    \item Classify the student’s step as \textit{Correct}, \textit{Suboptimal}, or \textit{Incorrect}.
    \item Provide brief, scaffolded feedback guiding the student toward the optimal step.
\end{enumerate}
\textbf{Constraints:}
\begin{itemize}[leftmargin=*, itemsep=0pt, topsep=0pt]
    \item Do not reveal the optimal step, its rule, or parent statements.
    \item Acknowledge what the student did correctly before addressing errors.
    \item Use Socratic questions to guide reasoning; keep feedback concise (2--3 sentences).
    \item Use predefined inference rule short names only.
\end{itemize}
\textbf{Response Format:}
\begin{itemize}[leftmargin=*, itemsep=0pt, topsep=0pt]
    \item \texttt{STUDENT\_ERRORS}: Brief explanation or \texttt{Correct}
    \item \texttt{NEXT\_STEP\_CORRECTNESS}: \texttt{Correct} / \texttt{Suboptimal} / \texttt{Incorrect}
    \item \texttt{TUTOR\_FEEDBACK}: Scaffolded guidance without answer revelation
\end{itemize}
\end{tcolorbox}
\caption{Base System prompt for Tutor}
\label{fig:tutor-prompt}
\end{figure}

\subsection{Teacher Prompt}
\label{appendix:teacher}

The Teacher agent has access to the complete next step and provides scaffolded feedback without revealing the answer. The prompt is illustrated in Fig \ref{fig:teacher-prompt}.

\begin{figure}[H]
\centering
\begin{tcolorbox}[
    colback=white,
    colframe=black,
    boxrule=0.5pt,
    arc=2pt,
    left=3pt,
    right=3pt,
    top=3pt,
    bottom=3pt,
    title=\textbf{Teacher Prompt},
    fonttitle=\bfseries,
]
\textbf{Role:} You are a Teacher evaluating a student’s proposed next step
in a propositional logic proof, with access to the complete solution
(\texttt{KNOWLEDGE\_BASE\_STEPS}).

\medskip
\textbf{Task:}
\begin{enumerate}[leftmargin=*, itemsep=0pt, topsep=0pt]
    \item Compare the student’s response against the knowledge-base solution.
    \item Identify errors in the student’s logic, rule usage, or reasoning.
    \item Classify the student’s next step as \textit{Correct}, \textit{Suboptimal}, or \textit{Incorrect}.
    \item Provide brief, scaffolded feedback guiding the student toward the correct solution.
\end{enumerate}
\textbf{Constraints:}
\begin{itemize}[leftmargin=*, itemsep=0pt, topsep=0pt]
    \item Do not reveal the exact next step, rule, or parent statements from the solution.
    \item Acknowledge correct aspects of the student’s attempt before addressing errors.
    \item Use Socratic questions to guide reasoning; keep feedback concise (2--3 sentences).
    \item Refer to the student’s candidates when relevant.
    \item Use predefined inference rule short names only.
\end{itemize}
\textbf{Response Format:}
\begin{itemize}[leftmargin=*, itemsep=0pt, topsep=0pt]
    \item \texttt{STUDENT\_ERRORS}: Brief explanation or \texttt{Correct}
    \item \texttt{NEXT\_STEP\_CORRECTNESS}: \texttt{Correct} / \texttt{Suboptimal} / \texttt{Incorrect}
    \item \texttt{TEACHER\_FEEDBACK}: Scaffolded guidance without answer revelation
\end{itemize}
\end{tcolorbox}
\caption{Base System prompt for Teacher}
\label{fig:teacher-prompt}
\end{figure}

\subsection{Judge Prompt}
\label{appendix:judge-direct}
The Judge agent is tasked with thoroughly evaluating both the student's and the teacher's responses to the logic proof problem. While the Teacher has access to the correct step and is therefore not asked to verify its own guidance, the Judge is specifically instructed to examine and identify any errors or inaccuracies present in both the student's and the teacher's responses.

Equipped with comprehensive access to the correct derivation steps from the knowledge base, the Judge carefully analyzes the reasoning behind both responses. Its role is to provide targeted evaluations and, where necessary, concise and actionable feedback to ensure alignment with the correct logic step. This approach ensures a robust layer of verification and strengthens the overall quality and pedagogical value of the feedback offered to the student. The prompt is illustrated in Fig \ref{fig:judge-prompt}.
\begin{figure}
\centering
\begin{tcolorbox}[
    colback=white,
    colframe=black,
    boxrule=0.5pt,
    arc=2pt,
    left=4pt,
    right=4pt,
    top=4pt,
    bottom=4pt,
    title=\textbf{Judge (Verifier) Prompt},
    fonttitle=\bfseries,
]

\textbf{Role:} You are an expert pedagogical AI Judge for propositional logic
proof problems, with access to the complete solution
(\texttt{KNOWLEDGE\_BASE\_STEPS}). You evaluate both the student’s proposed
next step and the Teacher’s feedback.
\medskip
\textbf{Task:}
\begin{enumerate}[leftmargin=*, itemsep=0pt, topsep=0pt]
    \item Compare the student’s response against the knowledge-base solution.
    \item Identify errors in the student’s reasoning, if any.
    \item Classify the student’s next step as \textit{Correct}, \textit{Suboptimal}, or \textit{Incorrect}.
    \item Evaluate whether the Teacher’s feedback correctly guides the student.
    \item Either enhance the Teacher’s feedback or override it with corrected guidance.
\end{enumerate}
\textbf{Constraints:}
\begin{itemize}[leftmargin=*, itemsep=0pt, topsep=0pt]
    \item Do not reveal the exact next step, rule, or parent statements from the solution.
    \item Acknowledge correct aspects of the student’s attempt before addressing errors.
    \item Use Socratic questions to guide reasoning; scaffold rather than instruct.
    \item Override Teacher feedback if it is incorrect, misleading, or reveals the solution.
    \item Keep final feedback concise (2--3 sentences) and encouraging.
    \item Use predefined inference rule short names only.
\end{itemize}
\textbf{Response Format:}
\begin{itemize}[leftmargin=*, itemsep=0pt, topsep=0pt]
    \item \texttt{STUDENT\_ERRORS}: Brief explanation or \texttt{Correct}
    \item \texttt{NEXT\_STEP\_CORRECTNESS}: \texttt{Correct} / \texttt{Suboptimal} / \texttt{Incorrect}
    \item \texttt{TEACHER\_FEEDBACK\_CORRECTNESS}: Assessment of Teacher feedback
    \item \texttt{JUDGE\_ACTION}: \texttt{Enhanced} or \texttt{Overridden}
    \item \texttt{FINAL\_FEEDBACK}: Judge-approved scaffolded guidance
\end{itemize}
\end{tcolorbox}
\caption{Base System prompt for Judge}
\label{fig:judge-prompt}
\end{figure}

\section{Illustrative Agent Responses}
\label{appendix:agent-responses}
This appendix presents representative responses generated by the Student, Tutor, Teacher, and Judge agents to concretely illustrate how each role operationalizes its assigned information access and reasoning constraints. These examples are intended to complement the quantitative analysis by providing qualitative insight into agent behavior across the tutoring pipeline. The responses are color-coded, \textcolor{red}{red} represents an incorrect student response or tutoring agent feedback, while \textcolor{blue}{blue} represents correct responses and feedbacks. 

\subsection{Student Agent Response}
We show a representative Student agent response to demonstrate candidate generation, explicit reasoning, and next-step generation. The example corresponds to a single proof instance and highlights how intermediate reasoning is externalized for downstream tutoring and verification. The sample response is shown in Fig \ref{fig:student-response}.

\begin{figure}[H]
\centering
\begin{tcolorbox}[
    colback=white,
    colframe=black,
    boxrule=0.5pt,
    arc=2pt,
    left=3pt,
    right=3pt,
    top=3pt,
    bottom=3pt,
    title=\textbf{Sample Instance},
    fonttitle=\bfseries
]

\textit{\textbf{Givens:}}
\begin{itemize}[leftmargin=*, itemsep=0pt, topsep=1pt]
    \item $((\neg K \lor L) \rightarrow (M \land N))$
    \item $(K \rightarrow O)$
    \item $\neg O$
\end{itemize}
\textit{\textbf{Intermediates:}}
\begin{enumerate}[leftmargin=*, itemsep=0pt, topsep=1pt]
    \item $\neg K$ \hfill (MT: 2,3)
    \item $(\neg K \lor L)$ \hfill (Add: 4)
    \item $(M \land N)$ \hfill (MP: 1,5)
\end{enumerate}
\textit{\textbf{Conclusion:}} $N$
\medskip
\tcbsubtitle{Corresponding student Response}
\textit{Candidate Steps:}
\begin{itemize}[leftmargin=*, itemsep=0pt, topsep=1pt]
    \item \textcolor{blue}{$M$: potentially useful for deriving $N$ from $(M \land N)$}.
    \item $\neg M$:  less direct without a clear derivation.
    \item $(M \land N)$:  already derived; insufficient alone to reach $N$.
\end{itemize}

\textit{Reasoning:  }
\textcolor{blue}{Applying a rule directly to $(M \land N)$ is the most efficient path to the conclusion. Since $(M \land N)$ is available, simplification yields the target.}
\\
\textit{Next Step:} \textcolor{blue}{$N$} \\
\textit{Rule:} \textcolor{blue}{Simp} \\
\textit{Parent Statement:} \textcolor{blue}{$(M \land N)$} \\
\end{tcolorbox}
\caption{Sample problem instance and corresponding \textcolor{blue}{\textbf{correct}} student response}
\label{fig:student-response}
\end{figure}

\subsection{Tutor Agent Response}
\label{appendix:tutor-response}
This example illustrates a Tutor agent interaction under \emph{partial solution access}. The Tutor receives only the next step and must provide scaffolded feedback without access to the full derivation context or rule justification. The sample response is shown in Fig \ref{fig: Tutor-response}.

\begin{tcolorbox}[
    colback=white,
    colframe=black,
    boxrule=0.5pt,
    arc=2pt,
    left=6pt,
    right=6pt,
    top=6pt,
    bottom=6pt,
    fonttitle=\bfseries,
    title={Tutor Response Interaction},
    breakable
]

\textbf{Problem Instance (Excerpt)}
\begin{itemize}[leftmargin=*, itemsep=1pt, topsep=2pt]
    \item $(Z \rightarrow (\neg Y \rightarrow X))$
    \item $(Z \land \neg W)$
    \item ...
\end{itemize}
\medskip
\textbf{Derived Intermediates (Excerpt)}
\begin{itemize}[leftmargin=*, itemsep=0pt, topsep=0pt]
    \item $\neg W$ \hfill (Simplification)
    \item $Z$ \hfill (Simplification)
    \item $(\neg Y \rightarrow X)$ \hfill (Modus Ponens)
    \item ...
\end{itemize}

\tcbsubtitle{Student Response}
\begin{itemize}[leftmargin=*, itemsep=0pt, topsep=0pt]
    \item \textbf{Candidates:}
    \begin{itemize}[leftmargin=1.2em, itemsep=1pt]
        \item $(Y \lor X)$: \textcolor{blue}{\textit{conjunction aiming to reach $(X \lor Y)$}}...
        \item ...
    \end{itemize}
    \item \textbf{Reasoning:}  
    \textcolor{blue}{\textit{$(X \lor Y)$} is optimal because it could be obtained via Addition from either $X$. }

    \item \textbf{Predicted Step:} \textcolor{blue}{$(X \lor Y)$}  
    \item \textbf{Rule:} Add  
    \item \textbf ...
\end{itemize}

\tcbsubtitle{Tutor Hint}
\begin{itemize}[leftmargin=*, itemsep=0pt, topsep=0pt]
    \item \textbf{Hint:} $(Y \lor X)$
\end{itemize}

\tcbsubtitle{Tutor Feedback}
\begin{itemize}[leftmargin=*, itemsep=1pt, topsep=2pt]
    \item \textbf{Tutor Rule: } Addition
    \item \textbf{Student Assessment:} incorrect
    \item \textbf{Feedback:} \textcolor{blue}{\emph{“You have correctly identified the next step...”}}
\end{itemize}
\end{tcolorbox}
\captionof{figure}{Tutor feedback (partial hint) on \textcolor{blue}{\textbf{correct} student response}} 
\label{fig: Tutor-response}

\subsection{Teacher Agent Response}
\label{appendix:teacher-response}
This example illustrates a Teacher agent with full derivational access, including the correct rule and parent statements. Unlike the Tutor, the Teacher can precisely diagnose suboptimal reasoning and scaffold guidance using explicit structural knowledge while avoiding direct answer revelation. The sample response is shown in Fig \ref{fig:teacher-response}.

\begin{tcolorbox}[
    colback=white,
    colframe=black,
    boxrule=0.5pt,
    arc=2pt,
    left=6pt,
    right=6pt,
    top=6pt,
    bottom=6pt,
    fonttitle=\bfseries,
    title={Teacher Response Interaction},
    breakable
]
\textbf{Problem Context (Excerpt)}
\begin{itemize}[leftmargin=*, itemsep=1pt, topsep=2pt]
    \item $(F \rightarrow (G \land \neg H)$
    \item $(I \land F)$
    \item $(H \lor J)$
    \item ...
\end{itemize}
\tcbsubtitle{Student Response}
\begin{itemize}[leftmargin=*, itemsep=1pt, topsep=2pt]
    \item \textbf{Candidates}
    \begin{itemize}
        \item \textcolor{blue}{(J+K) by Addition from J}
        \item ...
    \end{itemize}
    \item \textbf{Reasoning:} \textcolor{blue}{(J+K) is optimal because it directly derives ...}
    \item ...
\end{itemize}
\tcbsubtitle{Teacher Hint}
Derive (J+K) from J using the Addition rule.
\tcbsubtitle{Teacher Response}
\begin{itemize}[leftmargin=*, itemsep=1pt, topsep=2pt]
    \item \textbf{Next-Step Correctness:} Correct
    \item \textbf{Student Errors:}  None
    \item \textbf{Teacher Feedback: }\textcolor{blue}{\textit{You correctly identified that Addition on J directly achieves the conclusion...}}
\end{itemize}
\end{tcolorbox}
\captionof{figure}{Teacher feedback (complete hint) on \textcolor{blue}{\textbf{correct} student response}} 
\label{fig:teacher-response}

\begin{tcolorbox}[
    colback=white,
    colframe=black,
    boxrule=0.4pt,
    arc=2pt,
    left=4pt,
    right=4pt,
    top=4pt,
    bottom=4pt,
    breakable,
    title=\textbf{Teacher Reveals Solution},
    fonttitle=\bfseries
]

\textbf{Givens:}
\begin{itemize}[leftmargin=*, itemsep=0pt, topsep=2pt]
    \item $(S > D) \lor I$
    \item $(-S \lor Q) > Y$
    \item $\neg D$
    \item $\neg D > \neg I$
\end{itemize}

\medskip
\tcbsubtitle{Student Response}
\begin{itemize}[leftmargin=*, itemsep=0pt, topsep=2pt]
    \item \textbf{Reasoning:} \textcolor{red}{MT on $\neg D > \neg I$ and $\neg D$ is needed to get $\neg I$ to be used later with $(S > D) \lor I$ for DS}
    \item \textbf{Next Step:} \textcolor{red}{$\neg I$}
    \item \textbf{Rule Used:} \textcolor{red}{MT}
\end{itemize}
\tcbsubtitle{Knowledge-Graph Correct Hint:}
Derive $\neg S$ from $(S > D)$ and $\neg D$ using \emph{Modus Tollens (MT)}.

\tcbsubtitle{Teacher Response}
\textbf{Teacher Feedback:}  
\textcolor{red}{Great job identifying the relevant candidates! You correctly spotted that $(\neg D \rightarrow \neg I)$ and $\neg D$ could be useful, but there's a small mix-up with the rule. Recall: MT requires a negated *consequent*, while MP works with a *true antecedent*. How you might use MP on $(S > D)$ and $\neg D$ to get $\neg D$.}
\end{tcolorbox}
\captionof{figure}{Teacher (complete hint) feedback \textcolor{red}{reveals solution}  on \textcolor{red}{\textbf{incorrect}} student response}
\label{fig:teacher-reveals-response}

\subsection{Judge Verification Response}
\label{appendix:judge-response}
This example illustrates a complete tutoring interaction, showing how the student's reasoning is evaluated by a Tutor with partial solution access, refined by a Judge with full derivational access. The instance highlights how verification can override pedagogically suboptimal feedback when the tutor does not produce good feedback. The sample response is shown in Fig \ref{fig:tutor-echo-verification} and Fig \ref{fig:tutor-ambiguous-verification}.

\begin{tcolorbox}[
    colback=white,
    colframe=black,
    boxrule=0.5pt,
    arc=2pt,
    left=6pt,
    right=6pt,
    top=6pt,
    bottom=6pt,
    fonttitle=\bfseries,
    title={Judge Verification Improves Tutor Echo},
    breakable
]

\textbf{Problem Context (Excerpt)}
\begin{itemize}[leftmargin=*, itemsep=0pt, topsep=1pt]
    \item $(S \rightarrow D)$
    \item $((\neg S \lor Q) \rightarrow Y)$
    \item $\neg D$
    \item $(\neg D \rightarrow \neg I)$
    \item ...
\end{itemize}
\tcbsubtitle{Student Response}
\begin{itemize}[leftmargin=*, itemsep=0pt, topsep=1pt]
    \item \textbf{Candidates:}
    \begin{itemize}[leftmargin=*, itemsep=0pt]
        \item \textcolor{red}{Simplify $S \rightarrow D$ to get $\neg S$}
        \item ...
    \end{itemize}
    \item \textbf{Reasoning:} \textcolor{red}{\textit{$\neg S$ is optimal because we already have $(S \rightarrow D)$ from INTERMEDIATE STEPS, and simplifying it to $\neg S$ allows us to construct $(\neg S \lor Q)$...}}
    \item \textbf{Rule:} \textcolor{red}{Simp}
\end{itemize}
\tcbsubtitle{Tutor Hint}
$\neg S$
\tcbsubtitle{Tutor Response}
\begin{itemize}[leftmargin=*, itemsep=0pt, topsep=1pt]
    \item \textbf{Tutor Rule:} \textcolor{red}{Simp - \textit{echo failure}}
    \item \textbf{Student Assessment:} \textcolor{red}{Correct}
    \item \textbf{Tutor Feedback:}  
    \textcolor{red}{\textit{Excellent reasoning! You correctly identified how simplifying $(S \rightarrow D)$ helps set up the next step toward the conclusion...}}
\end{itemize}
\tcbsubtitle{Judge Verification}
\begin{itemize}[leftmargin=*, itemsep=0pt, topsep=1pt]
    \item \textbf{Tutor Feedback Correctness:} \textcolor{red}{Incorrect}
    \item \textbf{Judge Action:} \textcolor{red}{Overridden Teacher Feedback}
    \item \textbf{Final Judge Feedback:} \textcolor{blue}{\textit{Great job recognizing that you need -S to proceed! However, be careful with the rules: you can't directly use Simp on a conditional like $(S \rightarrow D)$, think about what rule you can use on $(S \rightarrow D)$ to turn it into a form more suitable for Simp...}}
\end{itemize}
\end{tcolorbox}
\captionof{figure}{Juge Verification Response, correcting \textcolor{red}{Incorrect} Tutor feedback (anchoring on \textcolor{red}{incorrect student reasoning)}, \textcolor{blue}{Judge verification overriding incorrect feedback.}}
\label{fig:tutor-echo-verification}

\begin{tcolorbox}[
    colback=white,
    colframe=black,
    boxrule=0.5pt,
    arc=2pt,
    left=4pt,
    right=4pt,
    top=4pt,
    bottom=4pt,
    title=\textbf{Judge verification helps Tutor Ambiguous Feedback},
    fonttitle=\bfseries,
    breakable
]
\textbf{Relevant Givens (subset):}
\[
((\neg Q + P) \rightarrow S), \quad (\neg T), \quad ((\neg T * S) \rightarrow \neg R)
\]

\tcbsubtitle{Student Response}
\begin{itemize}[leftmargin=*, itemsep=2pt]
    \item \textbf{Reasoning:}  
    \textcolor{red}{DS on $((\neg Q + P) \rightarrow S)$ and $(\neg Q + P)$ is optimal because it directly yields $S$, which can combine with $\neg T$ to form $(\neg T * S)$ and move toward deriving $\neg R$.}
    \item \textbf{Proposed Step:} \textcolor{red}{$S$}
    \item \textbf{Applied Rule:} \textcolor{red}{Disjunctive Syllogism (DS)}
\end{itemize}
\tcbsubtitle{Tutor Hint}
    $S$ 
\tcbsubtitle{Teacher Feedback}
\begin{itemize}[leftmargin=*, itemsep=2pt]
    \item \textbf{Tutor Rule:} \textcolor{red}{Disjunctive Syllogism (DS)} (echo failure)
    \item \textbf{Feedback Text:}  
    \textcolor{red}{\emph{Excellent reasoning! You’ve identified the most direct path to derive a key component.  
    Now, how might this newly derived statement interact with the other givens to progress toward the conclusion?}} (ambiguous and generic feedback)
\end{itemize}
\textbf{Judge (Verifier) Response}
\begin{itemize}[leftmargin=*, itemsep=2pt]
    \item \textbf{Teacher Feedback Correctness:} \textcolor{blue}{\textit{Vague}}
    \item \textbf{Judge Action:} \textcolor{blue}{\textit{Overridden teacher's feedback}}
    \item \textbf{Judge Feedback:}  
    \textcolor{blue}{\emph{Great job spotting that $S$ is a key component! You’re right that deriving $S$ brings you closer to using $((\neg T * S)\rightarrow \neg R)$. Now, consider: which rule actually applies to $((\neg Q + P)\rightarrow S)$ and $(\neg Q + P)$ to derive $S$? How might this step help you connect to the other givens, like $\neg T$?}}
\end{itemize}
\end{tcolorbox}
\captionof{figure}{Judge Verification Response, \textcolor{red}{Tutor echoes incorrect student rule and generates ambiguous and vague feedback} - \textcolor{blue}{Judge with full solution access helps}}
\label{fig:tutor-ambiguous-verification}

\section{Implementation Details}
\label{appendix:implementation}

\paragraph{Response Constraints.}
We standardize inference rule references using abbreviations (MP, MT, DS, 
HS, DeM, Impl) for consistent evaluation. The complete list appears in 
Table \ref{tab:inference-rules}.

\paragraph{Token Limits.}
Rather than hard limits, we employ soft constraints via prompts (2--3 sentences). 

\paragraph{Quality Validation.}
Two automated mechanisms ensure data integrity: (1) zero-token outputs 
trigger automatic retry, and (2) responses missing required JSON fields 
or violating constraints trigger re-prompting. Persistent failures after 
three attempts are flagged for manual review. Across 516 instances on 7 
models, fewer than 3\% required retry.

\section{Model Specifications}
\label{appendix:model-specs}

\begin{table}[H]
\centering
\caption{Models used in our experiments}
\label{tab:models}
\footnotesize
\small
\setlength{\tabcolsep}{4pt}  
\begin{tabular}{l c c c c}
\hline
\textbf{Model} &
{\scriptsize\textbf{Org.}} &
{\scriptsize\textbf{API}} &
{\scriptsize\textbf{Context}} &
{\scriptsize\textbf{Param}} \\
\hline
GPT-4.1 & OpenAI & OpenAI & 1M &
- \\
GPT-o3 & OpenAI & OpenAI & 1M &
- \\
Gemini 1.5 Pro & Google & Vertex & 128K &
- \\
Qwen-3-32B & Alibaba & Groq & 32K &
32B \\
LLaMA-3.3-70B & Meta & Groq & 128K &
70B \\
Mistral Large & Mistral & Mistral & 256K &
675B \\
Deepseek-r1-0528 & Deepseek & Deepinfra & 128K &
671B \\
\hline
\end{tabular}
\caption*{\footnotesize\RaggedRight"\--" refers to undisclosed   
}
\end{table}

\section{Manual Evaluation Protocol}
\label{appendix:human-eval}

To complement our automated metrics and better understand qualitative differences in feedback behavior across pipelines, we conducted a manual evaluation of a representative subset of model responses. This analysis focuses on pedagogical properties of feedback that are difficult to capture through accuracy-based metrics alone, such as scaffolding quality and answer revelation. Three domain experts independently annotated 10 problem instances per feedback pipeline (Student, Tutor, Teacher, and Judge), resulting in a total of 40 responses per model along four pedagogical dimensions. We use a minimally sufficient sampling strategy until inter-rater agreement exceeded a moderate Krippendorff’s $\alpha$ threshold ($\alpha \geq 0.65$).  

\subsection{Annotators}
Three expert annotators participated in the study. All annotators had at least two years of experience as teaching assistants for undergraduate Discrete Structures and had previously served as researchers for the logic tutoring system used to generate the dataset. This ensured familiarity with both propositional logic proof strategies and pedagogical objectives of step-level tutoring.

\subsection{Sample Selection}
We randomly sampled 10 problem instances per feedback pipeline (Tutor, Teacher, and Judge), resulting in a total of 40 annotated responses per model. Sampling includes a mix of correct, suboptimal, and incorrect student attempts. Following best practices for qualitative analysis in educational NLP, we used a minimally sufficient sampling strategy, increasing the sample size until inter-annotator agreement reached a stable, moderate level, Krippendorff’s $\alpha \geq 0.70$ across all annotation dimensions.

\subsection{Annotation Dimensions}
Each feedback response was independently annotated along four pedagogically motivated dimensions:

\paragraph{Evaluation Rubrics.}
Feedback was annotated along four dimensions capturing both reasoning accuracy
and pedagogical effectiveness:

\begin{itemize}[leftmargin=*, itemsep=2pt]
    \item \textbf{Mistake Identification}: Whether the feedback correctly
    identifies errors or suboptimality in the student’s proposed next step,
    including incorrect rule use, invalid derivations, or deviations from the
    optimal reasoning path.

    \item \textbf{Correctness}: Whether the feedback itself is logically sound
    and consistent with the correct solution, providing guidance that does not
    introduce new errors or misleading reasoning.

    \item \textbf{Answer Revelation}: Whether the feedback explicitly or
    implicitly reveals the correct next step, inference rule, or parent
    statements, thereby reducing the need for student reasoning.

    \item \textbf{Actionability}: Whether the feedback provides clear, concrete,
    and usable guidance that a student could plausibly act on to improve their
    next attempt.
\end{itemize}

Due to the cognitive complexity of step-level logic feedback and the time cost of expert annotation, we adopt a coarse 3-point ordinal scale (1–3) for all rubrics, with 1 = poor, 2 = partial, 3 = strong. This scale captures meaningful distinctions while maintaining annotation reliability and consistency.

\subsection{Annotation Procedure}
Annotators were provided with the problem context (givens, intermediates, and conclusion), the student’s response, and the corresponding feedback generated by the pipeline under evaluation. Annotators worked independently and were blinded to model identity and quantitative performance results. 
Inter-annotator agreement was measured using Krippendorff’s $\alpha$, which is suitable for ordinal data and multiple annotators. Agreement exceeded a moderate reliability threshold across all four dimensions ($\alpha \geq 0.65 $), indicating consistent interpretation of the rubric.

\subsection{Analysis}
Although qualitative and limited in scope, this analysis reveals consistent patterns that align with our quantitative results. Tutor feedback is typically vague but provides scaffolding, encouraging reasoning without explicit solution disclosure, whereas Teacher feedback often encodes the correct inference more directly (e.g., naming rules or premises). Judge feedback, when applied to noisy or overly revealing Teacher responses, tends to over-correct or abstract guidance, reducing instructional specificity, mirroring the conditional effectiveness of verification observed in our experiments.

\section{Design Implications for Education}
\label{sec:design-implications}
Our findings have direct implications for the design of LLM-based tutoring systems in formal reasoning domains. First, verification should be conditionally deployed, not treated as a universal safeguard. Systems should estimate upstream tutor reliability and invoke verification primarily when error rates are high, rather than indiscriminately. Second, solution information access must be adaptive. Full derivational access expands coverage to harder states but increases the risk of premature revelation; lightweight access is preferable in low-complexity settings. Third, step-level reasoning alignment is critical: feedback that is locally valid but misaligned with the learner’s current proof state can anchor incorrect reasoning and undermine learning. Finally, our results reveal a complexity ceiling beyond which no automated feedback strategy is reliable, suggesting the need for escalation strategies such as human intervention or extended symbolic reasoning support in highly complex cases.

\section{Future Directions}
\label{sec:future}
Several directions remain for future research. First, validating these findings with human learners is essential to assess learning gains, engagement, and transfer beyond LLM-based student simulations. Second, extending the framework to multi-turn tutoring dialogues may reveal different verification dynamics, including delayed correction or cumulative anchoring effects. Third, applying our approach to broader formal reasoning domains, such as first-order logic or program verification, would test its generality. Finally, developing adaptive routing mechanisms that dynamically select tutor roles, verification depth, and solution access based on student state and problem difficulty represents a promising direction toward robust, scalable educational AI systems.

\setlength{\floatsep}{6pt}
\setlength{\textfloatsep}{6pt}
\setlength{\intextsep}{6pt}

\begin{table*}[t]
\centering
\small
\caption{Effects of proof complexity on correctness and feedback pipelines (Logistic Regression Coefficients, significance: *** $p<.001$, ** $p<.01$, * $p<.05$)}
\label{tab:logit-effects}
\begin{tabular}{lcccc}
\toprule
\textbf{Model} &
\multicolumn{1}{c}{\textbf{Main Effect}} &
\multicolumn{3}{c}{\textbf{Interaction with Pipeline}} \\
\cmidrule(lr){2-2}
\cmidrule(lr){3-5}
&
\textbf{Complexity} &
\textbf{Tutor $\times$ Comp.} &
\textbf{Teacher $\times$ Comp.} &
\textbf{Judge $\times$ Comp.} 
\\
\midrule
GPT-4.1      & $-0.35^{***}$ & $-0.02^{ns}$ & $+0.08^{ns}$ & $+0.22^{***}$ \\
GPT-o3       & $-0.42^{***}$ & $+0.18^{*}$ & $+0.13^{ns}$ & $+0.24^{**}$  \\
Gemini-1.5   & $-0.55^{***}$ & $+0.55^{***}$ & $+0.65^{***}$ & $+0.61^{***}$  \\
DeepSeek-R1  & $-0.36^{***}$ & $+0.28^{***}$ & $+0.36^{***}$ & $+0.56^{***}$  \\
Qwen-3       & $-0.53^{***}$ & $+0.33^{***}$ & $+0.37^{***}$ & $+0.39^{***}$  \\
Mistral      & $-0.42^{***}$ & $+0.10^{ns}$ & $+0.11^{ns}$ & $+0.21^{*}$ \\
LLaMA-3      & $-0.30^{***}$ & $-0.03^{ns}$ & $-0.03^{ns}$ & $+0.04^{ns}$  \\
\bottomrule
\end{tabular}
\end{table*}

\begin{table*}[t]
\centering
\small
\caption{Feedback hurts student performance with corresponding mean complexity}
\label{tab:degrade}
\begin{tabular}{lcccccc}
\toprule
\textbf{Model} &
\multicolumn{2}{c}{\textbf{Tutor}} &
\multicolumn{2}{c}{\textbf{Teacher}} &
\multicolumn{2}{c}{\textbf{Judge}} \\
\cmidrule(lr){2-3} \cmidrule(lr){4-5} \cmidrule(lr){6-7}
 & \textbf{Count} & \textbf{Complexity} 
 & \textbf{Count} & \textbf{Complexity} 
 & \textbf{Count} & \textbf{Complexity} \\
\midrule
GPT        & 295 & 3.97 & 200 & 4.61 & 127 & 4.90 \\
GPT-o3     & 267 & 4.14 & 200 & 4.64 & 126 & 5.23 \\
Gemini     & 134 & 4.79 & 109 & 5.73 & 139 & 4.82 \\
DeepSeek   & 121 & 4.74 &  81 & 6.12 & 104 & 5.08 \\
LLaMA      & 383 & 3.44 & 332 & 3.60 & 307 & 3.60 \\
Qwen       & 218 & 4.50 & 200 & 4.74 & 182 & 4.74 \\
Mistral    & 319 & 3.84 & 302 & 3.94 & 226 & 4.23 \\
\bottomrule
\end{tabular}
\end{table*}

\begin{table*}[t]
\centering
\small
\caption{Tutor rule inference accuracy by rule type and model (\%), best values are bolded across all rows}
\begin{tabular}{llclccccc}
\toprule
\textbf{Rule} &  \textbf{Actual}&\textbf{GPT-4.1}&  \textbf{GPT-o3}&\textbf{Gemini} & \textbf{Mistral} & \textbf{Deepseek}& \textbf{Qwen} & \textbf{Llama-3.3}\\
\midrule
\textbf{Simp}&  90&     87&     86&    94&     73&     \textbf{96}&     76&     18 \\
\textbf{Add}&  61&      81&     77&    78&     62&     85&     \textbf{92}&      23\\
\textbf{MP}&   61&      70&     77&    \textbf{96}&     45&     91&     75&      55\\
\textbf{DS}&   53&      43&      45&    81&    60&     77&     \textbf{90}&      17\\
\textbf{MT}&   40&      50&     55&    \textbf{82}&     70&     87&     77&      10\\
\textbf{Impl}&  38&     47&     52&    \textbf{86}&     65&     81&     54&      21\\
\textbf{CP}&    36&     30&     33&    83&     10&     \textbf{86}&     58&      17\\
\textbf{DN}&    31&     48&     58&    \textbf{83}&     38&     74&      12&     29\\
\textbf{ CD}&   27&     7&      11&    \textbf{66}&     3&      \textbf{66}&      55&     11\\
\textbf{ DeM}&  21&     47&     38&    \textbf{95}&     66&     85&      66&     14\\
\textbf{ HS}&  21&      47&     57&    \textbf{100}&    80&     92&      85&     61\\
\textbf{ Conj}&  14&    \textbf{92}&     \textbf{92}&     \textbf{92}&    \textbf{92}&     83&      85&     0\\
\textbf{ Comm}&  12&    41&     33&    \textbf{72}&      14&    0&       8&      50\\
\textbf{ Equiv}&  7&    0&      0&      33&     \textbf{100}&   \textbf{100}&     0&       0\\
\textbf{ Distr}&  4&    \textbf{100}&    \textbf{100}&    \textbf{100}&      0&    0&      50&      50\\
\midrule
\textbf{Overall} &  &   58.3&     60.1&       \textbf{85.8}&     54.8&      \textbf{84.9}&   70.2 &    24       \\
\bottomrule
\end{tabular}

\label{tab:rule-acc-full}
\end{table*}

%% file: custom.bib
@article{macina2025mathtutorbench,
  title={Mathtutorbench: A benchmark for measuring open-ended pedagogical capabilities of llm tutors},
  author={Macina, Jakub and Daheim, Nico and Hakimi, Ido and Kapur, Manu and Gurevych, Iryna and Sachan, Mrinmaya},
  journal={arXiv preprint arXiv:2502.18940},
  year={2025}
}

@inproceedings{daheim2024stepwise,
  title={Stepwise Verification and Remediation of Student Reasoning Errors with Large Language Model Tutors},
  author={Daheim, Nico and Macina, Jakub and Kapur, Manu and Gurevych, Iryna and Sachan, Mrinmaya},
  booktitle={Proceedings of the 2024 Conference on Empirical Methods in Natural Language Processing},
  pages={8386--8411},
  year={2024}
}

@article{dos2025intelligent,
  title={INTELLIGENT MULTI-AGENT SYSTEMS WITH GENERATIVE AI IN EDUCATION: A SYSTEMATIC LITERATURE MAPPING},
  author={dos Santos, O Lube and Cury, D},
  journal={ICERI2025 Proceedings},
  pages={9037--9046},
  year={2025},
  publisher={IATED}
}

@inproceedings{phung2024automating,
  title={Automating human tutor-style programming feedback: Leveraging gpt-4 tutor model for hint generation and gpt-3.5 student model for hint validation},
  author={Phung, Tung and P{\u{a}}durean, Victor-Alexandru and Singh, Anjali and Brooks, Christopher and Cambronero, Jos{\'e} and Gulwani, Sumit and Singla, Adish and Soares, Gustavo},
  booktitle={Proceedings of the 14th learning analytics and knowledge conference},
  pages={12--23},
  year={2024}
}

@inproceedings{macina2023mathdial,
  title={Mathdial: A dialogue tutoring dataset with rich pedagogical properties grounded in math reasoning problems},
  author={Macina, Jakub and Daheim, Nico and Chowdhury, Sankalan and Sinha, Tanmay and Kapur, Manu and Gurevych, Iryna and Sachan, Mrinmaya},
  booktitle={Findings of the Association for Computational Linguistics: EMNLP 2023},
  pages={5602--5621},
  year={2023}
}

@inproceedings{chudziak2025ai,
  title={Ai-powered math tutoring: Platform for personalized and adaptive education},
  author={Chudziak, Jaros{\l}aw A and Kostka, Adam},
  booktitle={International Conference on Artificial Intelligence in Education},
  pages={462--469},
  year={2025},
  organization={Springer}
}

@inproceedings{shi-etal-2025-educationq,
    title = "{E}ducation{Q}: Evaluating {LLM}s' Teaching Capabilities Through Multi-Agent Dialogue Framework",
    author = "Shi, Yao  and
      Liang, Rongkeng  and
      Xu, Yong",
    editor = "Che, Wanxiang  and
      Nabende, Joyce  and
      Shutova, Ekaterina  and
      Pilehvar, Mohammad Taher",
    booktitle = "Proceedings of the 63rd Annual Meeting of the Association for Computational Linguistics (Volume 1: Long Papers)",
    month = jul,
    year = "2025",
    address = "Vienna, Austria",
    publisher = "Association for Computational Linguistics",
    url = "https://aclanthology.org/2025.acl-long.1576/",
    doi = "10.18653/v1/2025.acl-long.1576",
    pages = "32799--32828",
    ISBN = "979-8-89176-251-0",
}

@article{zhang2025eduplanner,
  title={Eduplanner: Llm-based multi-agent systems for customized and intelligent instructional design},
  author={Zhang, Xueqiao and Zhang, Chao and Sun, Jianwen and Xiao, Jun and Yang, Yi and Luo, Yawei},
  journal={IEEE Transactions on Learning Technologies},
  year={2025},
  publisher={IEEE}
}

@article{inamdar2025logiclearner,
  title={LogicLearner: A Tool for the Guided Practice of Propositional Logic Proofs},
  author={Inamdar, Amogh and Macar, Uzay and Vazirani, Michel and Tarnow, Michael and Mustapha, Zarina and Dittren, Natalia and Sadeh, Sam and Verma, Nakul and Salleb-Aouissi, Ansaf},
  journal={arXiv preprint arXiv:2503.19280},
  year={2025}
}

@inproceedings{barnes2008toward,
  title={Toward automatic hint generation for logic proof tutoring using historical student data},
  author={Barnes, Tiffany and Stamper, John},
  booktitle={International conference on intelligent tutoring systems},
  pages={373--382},
  year={2008},
  organization={Springer}
}

@inproceedings{huang2024selfcorrect,
  title     = {Large Language Models Cannot Self-Correct Reasoning Yet},
  author    = {Huang, Jie and others},
  booktitle = {International Conference on Learning Representations (ICLR)},
  year      = {2024}
}

@inproceedings{zhang2024small,
  title     = {Small Language Models Need Strong Verifiers to Self-Correct Reasoning},
  author    = {Zhang, Yifan and others},
  booktitle = {Findings of the Association for Computational Linguistics (ACL)},
  year      = {2024}
}

@book{vygotsky1978mind,
  title={Mind in society: Development of higher psychological processes},
  author={Vygotsky, Lev Semenovich and Cole, Michael},
  year={1978},
  publisher={Harvard university press}
}

@article{wood1976scaffolding,
  title={The Role of Tutoring in Problem Solving},
  author={Wood, David and Bruner, Jerome S and Ross, Gail},
  journal={Journal of Child Psychology and Psychiatry},
  volume={17},
  number={2},
  pages={89--100},
  year={1976},
  publisher={Wiley},
  note={Foundational work on scaffolding and just-in-time interventions}
}

@article{belland2017scaffolding,
  title={Instructional Scaffolding in STEM Education: Strategies and Efficacy Evidence},
  author={Belland, Brian R},
  journal={Springer Nature},
  year={2017},
  note={Comprehensive review of scaffolding strategies in STEM}
}

@article{vanlehn2011effectiveness,
  title={The Relative Effectiveness of Human Tutoring, Intelligent Tutoring Systems, and Other Tutoring Systems},
  author={VanLehn, Kurt},
  journal={Educational Psychologist},
  volume={46},
  number={4},
  pages={197--221},
  year={2011},
  publisher={Taylor \& Francis},
  note={Step-based ITS achieve effect sizes of 0.75-0.80, comparable to human tutoring (0.79)}
}

@article{mayer2025importance,
  title={The Importance of Teaching Logic to Computer Scientists and Electrical Engineers},
  author={Mayer, Paul and Baraniuk, Rich},
  journal={ACM Transactions on Computing Education},
  volume={25},
  number={2},
  pages={1--12},
  year={2025},
  publisher={ACM New York, NY}
}

@inproceedings{serban2020korbit,
  title={Automated Personalized Feedback Improves Learning Gains in an Intelligent Tutoring System},
  author={Serban, Iulian Vlad and others},
  booktitle={International Conference on Artificial Intelligence in Education},
  pages={140--146},
  year={2020},
  publisher={Springer},
  note={Personalized hints improve learning gains; feedback from ZPD}
}

@article{kochmar2022automated,
  title={Automated data-driven generation of personalized pedagogical interventions in intelligent tutoring systems},
  author={Kochmar, Ekaterina and Vu, Dung Do and Belfer, Robert and Gupta, Varun and Serban, Iulian Vlad and Pineau, Joelle},
  journal={International Journal of Artificial Intelligence in Education},
  volume={32},
  number={2},
  pages={323--349},
  year={2022},
  publisher={Springer}
}

@article{vanpol2010scaffolding,
  title={Scaffolding in Teacher–Student Interaction: A Decade of Research},
  author={Van de Pol, Janneke and Volman, Monique and Beishuizen, Jos},
  journal={Educational Psychology Review},
  volume={22},
  number={3},
  pages={271--296},
  year={2010},
  publisher={Springer},
  note={Comprehensive review of scaffolding in teacher-student interactions}
}

@article{chen2025multi,
  title={Multi-agent-as-judge: Aligning llm-agent-based automated evaluation with multi-dimensional human evaluation},
  author={Chen, Jiaju and Lu, Yuxuan and Wang, Xiaojie and Zeng, Huimin and Huang, Jing and Gesi, Jiri and Xu, Ying and Yao, Bingsheng and Wang, Dakuo},
  journal={arXiv preprint arXiv:2507.21028},
  year={2025}
}

@article{gonnermann2025facet,
  title={FACET: Teacher-Centred LLM-Based Multi-Agent Systems-Towards Personalized Educational Worksheets},
  author={Gonnermann-M{\"u}ller, Jana and Haase, Jennifer and Fackeldey, Konstantin and Pokutta, Sebastian},
  journal={arXiv preprint arXiv:2508.11401},
  year={2025}
}

@inproceedings{corbett2001cognitive,
  title={Cognitive computer tutors: Solving the two-sigma problem},
  author={Corbett, Albert},
  booktitle={International Conference on User Modeling},
  pages={137--147},
  year={2001},
  organization={Springer}
}

@article{chi2014icap,
  title={The ICAP framework: Linking cognitive engagement to active learning outcomes},
  author={Chi, Michelene TH and Wylie, Ruth},
  journal={Educational psychologist},
  volume={49},
  number={4},
  pages={219--243},
  year={2014},
  publisher={Taylor \& Francis}
}

@article{zerkouk2025its_review,
  title={A Comprehensive Review of AI-based Intelligent Tutoring Systems: Applications and Challenges},
  author={Zerkouk, Mohamed and Chikhaoui, Belkacem},
  journal={arXiv preprint arXiv:2507.18882},
  year={2025},
  note={ITS can improve student performance; CLT informs stepwise hint design}
}

@article{chen2023its_sustainable,
  title={Artificial Intelligence in Intelligent Tutoring Systems Toward Sustainable Education: A Systematic Review},
  author={Chen, others},
  journal={Smart Learning Environments},
  volume={10},
  pages={41},
  year={2023},
  publisher={Springer},
  note={ITS monitor progress, provide feedback, hints, and scaffolding}
}

@inproceedings{tafjord2021proofwriter,
  title={Proofwriter: Generating implications, proofs, and abductive statements over natural language},
  author={Tafjord, Oyvind and Dalvi, Bhavana and Clark, Peter},
  booktitle={Findings of the Association for Computational Linguistics: ACL-IJCNLP 2021},
  pages={3621--3634},
  year={2021}
}

@inproceedings{dhuliawala2024chain,
  title={Chain-of-verification reduces hallucination in large language models},
  author={Dhuliawala, Shehzaad and Komeili, Mojtaba and Xu, Jing and Raileanu, Roberta and Li, Xian and Celikyilmaz, Asli and Weston, Jason},
  booktitle={Findings of the Association for Computational Linguistics: ACL 2024},
  pages={3563--3578},
  year={2024}
}

@inproceedings{jia2024assessing,
  title={On assessing the faithfulness of llm-generated feedback on student assignments},
  author={Jia, Qinjin and Cui, Jialin and Xi, Ruijie and Liu, Chengyuan and Rashid, Parvez and Li, Ruochi and Gehringer, Edward},
  booktitle={Proceedings of the 17th International Conference on Educational Data Mining},
  pages={491--499},
  year={2024}
}

@inproceedings{liu2025improving,
  title={Improving AI in CS50: Leveraging Human Feedback for Better Learning},
  author={Liu, Rongxin and Zhao, Julianna and Xu, Benjamin and Perez, Christopher and Zhukovets, Yuliia and Malan, David J},
  booktitle={Proceedings of the 56th ACM Technical Symposium on Computer Science Education V. 1},
  pages={715--721},
  year={2025}
}

@inproceedings{scholz2025partnering,
  title={Partnering with AI: A pedagogical feedback system for LLM integration into programming education},
  author={Scholz, Niklas and Nguyen, Manh Hung and Singla, Adish and Nagashima, Tomohiro},
  booktitle={European Conference on Technology Enhanced Learning},
  pages={243--248},
  year={2025},
  organization={Springer}
}

@article{Kumar2024Supporting,
title={Supporting Self-Reflection at Scale with Large Language Models: Insights from Randomized Field Experiments in Classrooms},author={Harsh Kumar and Ruiwei Xiao and Benjamin Lawson and Ilya Musabirov and Jiakai Shi and Xinyuan Wang and Huayin Luo and J. Williams and Anna N. Rafferty and John Stamper and Michael Liut},journal={Proceedings of the Eleventh ACM Conference on Learning @ Scale},year={2024},doi={10.1145/3657604.3662042}}

@article{saparov2022language,
  title={Language models are greedy reasoners: A systematic formal analysis of chain-of-thought},
  author={Saparov, Abulhair and He, He},
  journal={arXiv preprint arXiv:2210.01240},
  year={2022}
}

@inproceedings{han2024folio,
  title={Folio: Natural language reasoning with first-order logic},
  author={Han, Simeng and Schoelkopf, Hailey and Zhao, Yilun and Qi, Zhenting and Riddell, Martin and Zhou, Wenfei and Coady, James and Peng, David and Qiao, Yujie and Benson, Luke and others},
  booktitle={Proceedings of the 2024 Conference on Empirical Methods in Natural Language Processing},
  pages={22017--22031},
  year={2024}
}

@article{reddig2025generating,
  title={Generating in-context, personalized feedback for intelligent tutors with large language models},
  author={Reddig, Jennifer M and Arora, Arav and MacLellan, Christopher J},
  journal={International Journal of Artificial Intelligence in Education},
  pages={1--42},
  year={2025},
  publisher={Springer}
}

@article{jiang2024llms,
  title={Llms can find mathematical reasoning mistakes by pedagogical chain-of-thought},
  author={Jiang, Zhuoxuan and Peng, Haoyuan and Feng, Shanshan and Li, Fan and Li, Dongsheng},
  journal={arXiv preprint arXiv:2405.06705},
  year={2024}
}

@article{zhou2022least,
  title={Least-to-most prompting enables complex reasoning in large language models},
  author={Zhou, Denny and Sch{\"a}rli, Nathanael and Hou, Le and Wei, Jason and Scales, Nathan and Wang, Xuezhi and Schuurmans, Dale and Cui, Claire and Bousquet, Olivier and Le, Quoc and others},
  journal={arXiv preprint arXiv:2205.10625},
  year={2022}
}

@inproceedings{srivatsa-etal-2025-llms-spot,
    title = "{LLM}s cannot spot math errors, even when allowed to peek into the solution",
    author = "Srivatsa, Kv Aditya  and
      Maurya, Kaushal Kumar  and
      Kochmar, Ekaterina",
    editor = "Christodoulopoulos, Christos  and
      Chakraborty, Tanmoy  and
      Rose, Carolyn  and
      Peng, Violet",
    booktitle = "Proceedings of the 2025 Conference on Empirical Methods in Natural Language Processing",
    month = nov,
    year = "2025",
    address = "Suzhou, China",
    publisher = "Association for Computational Linguistics",
    url = "https://aclanthology.org/2025.emnlp-main.553/",
    doi = "10.18653/v1/2025.emnlp-main.553",
    pages = "10914--10928",
    ISBN = "979-8-89176-332-6"
}

@misc{tyen2024llmsreasoningerrorscorrect,
      title={LLMs cannot find reasoning errors, but can correct them given the error location}, 
      author={Gladys Tyen and Hassan Mansoor and Victor Cărbune and Peter Chen and Tony Mak},
      year={2024},
      eprint={2311.08516},
      archivePrefix={arXiv},
      primaryClass={cs.AI},
      url={https://arxiv.org/abs/2311.08516}, 
}

@article{estevez2025evaluation,
  title={Evaluation of LLM tools for feedback generation in a course on concurrent programming},
  author={Est{\'e}vez-Ayres, Iria and Callejo, Patricia and Hombrados-Herrera, Miguel {\'A}ngel and Alario-Hoyos, Carlos and Delgado Kloos, Carlos},
  journal={International journal of artificial intelligence in education},
  volume={35},
  number={2},
  pages={774--790},
  year={2025},
  publisher={Springer}
}

@article{kamoi2024can,
  title={When can llms actually correct their own mistakes? a critical survey of self-correction of llms},
  author={Kamoi, Ryo and Zhang, Yusen and Zhang, Nan and Han, Jiawei and Zhang, Rui},
  journal={Transactions of the Association for Computational Linguistics},
  volume={12},
  pages={1417--1440},
  year={2024}
}
